\documentclass[runningheads]{llncs}
\usepackage{graphicx}
\usepackage{tikz}
\usepackage{comment}
\usepackage{amsmath,amssymb}
\usepackage{color}
\usepackage{algorithm,algorithmic}
\usepackage{booktabs}
\usepackage[T1]{fontenc}
\usepackage{mathrsfs}
\usepackage{amsfonts}
\usepackage{multirow}
\usepackage{diagbox}
\usepackage{makecell}
\usepackage{subcaption}
\usepackage{caption}
\usepackage{marvosym}
\usepackage[pagebackref,breaklinks,colorlinks]{hyperref}
\usepackage[accsupp]{axessibility}

\newcommand{\etal}{\textit{et al}.}
\newcommand{\ie}{\textit{i}.\textit{e}.}
\newcommand{\eg}{\textit{e}.\textit{g}.}
\newcommand{\etc}{\textit{etc}.}
\newcommand\blfootnote[1]{
  \begingroup
  \renewcommand\thefootnote{}\footnote{#1}
  \addtocounter{footnote}{-1}
  \endgroup
}

\usepackage[capitalize]{cleveref}
\crefname{section}{Sec.}{Secs.}
\Crefname{section}{Section}{Sections}
\Crefname{table}{Table}{Tables}
\crefname{table}{Tab.}{Tabs.}

\begin{document}
\pagestyle{headings}
\mainmatter

\title{Joint-Modal Label Denoising for Weakly-Supervised Audio-Visual Video Parsing}
\titlerunning{Joint-Modal Label Denoising for Weakly-Supervised AVVP}
\author{Haoyue Cheng$^{1,2}$ \,
Zhaoyang Liu$^{2}$ \,
Hang Zhou$^{3}$ \, Chen Qian$^{2}$ \\ Wayne Wu$^{2}$ \, Limin Wang$^{1,4}$\textsuperscript{\Letter}}
\authorrunning{H. Cheng, Z. Liu, H. Zhou, C. Qian, W. Wu, L. Wang}
\institute{State Key Laboratory for Novel Software Technology, Nanjing University, China \and SenseTime Research \ \ \
$^{3}$ CUHK - Sensetime Joint Lab \ \ \
$^{4}$ Shanghai AI Laboratory \\
\email{chenghaoyue98@gmail.com \ \ zyliumy@gmail.com \ \ zhouhang@link.cuhk.edu.hk \ \ qianchen@sensetime.com \ \ wuwenyan0503@gmail.com \ \ lmwang@nju.edu.cn}}
\maketitle

\begin{abstract}
This paper focuses on the weakly-supervised audio-visual video parsing task, which aims to recognize all events belonging to each modality and localize their temporal boundaries. This task is challenging because only overall labels indicating the video events are provided for training.
However, an event might be labeled but not appear in one of the modalities, which results in a modality-specific noisy label problem.
In this work, we propose a training strategy to identify and remove modality-specific noisy labels dynamically. It is motivated by two key observations: 1) networks tend to learn clean samples first; and 2) a labeled event would appear in at least one modality. 
%
Specifically, we sort the losses of all instances within a mini-batch individually in each modality, and
then select noisy samples according to the relationships between intra-modal and inter-modal losses. 
Besides, we also propose a simple but valid noise ratio estimation method by calculating the proportion of instances whose confidence is below a preset threshold.
Our method makes large improvements over the previous state of the arts (\eg, from 60.0\% to 63.8\% in segment-level visual metric), which demonstrates the effectiveness of our approach. Code and trained models are publicly available
at \url{https://github.com/MCG-NJU/JoMoLD}.
\end{abstract}
\blfootnote{\small \Letter: Corresponding author.}

\section{Introduction}

Many works show that the audio-visual clues play a crucial role in comprehensive video understanding~\cite{arandjelovic2017look,xiao2020audiovisual,panda2021adamml}.
However, most studies on audio-visual joint learning~\cite{tian2018audio,hu2019deep,morgado2021avid} assume that the two modalities are correlated or even temporally synchronized, which is not always the case.
For example, the sound of a car might be out of sight, but such information is still crucial for real-world perception.
To this end, Tian \etal~\cite{tian2020unified} proposed the \textbf{audio-visual video parsing} (AVVP) task without consistency restriction, which aims to recognize all events belonging to each modality and localize their temporal boundaries.

\begin{figure}[t]
  \centering
  \includegraphics[width=0.8\textwidth,height=4.5cm]{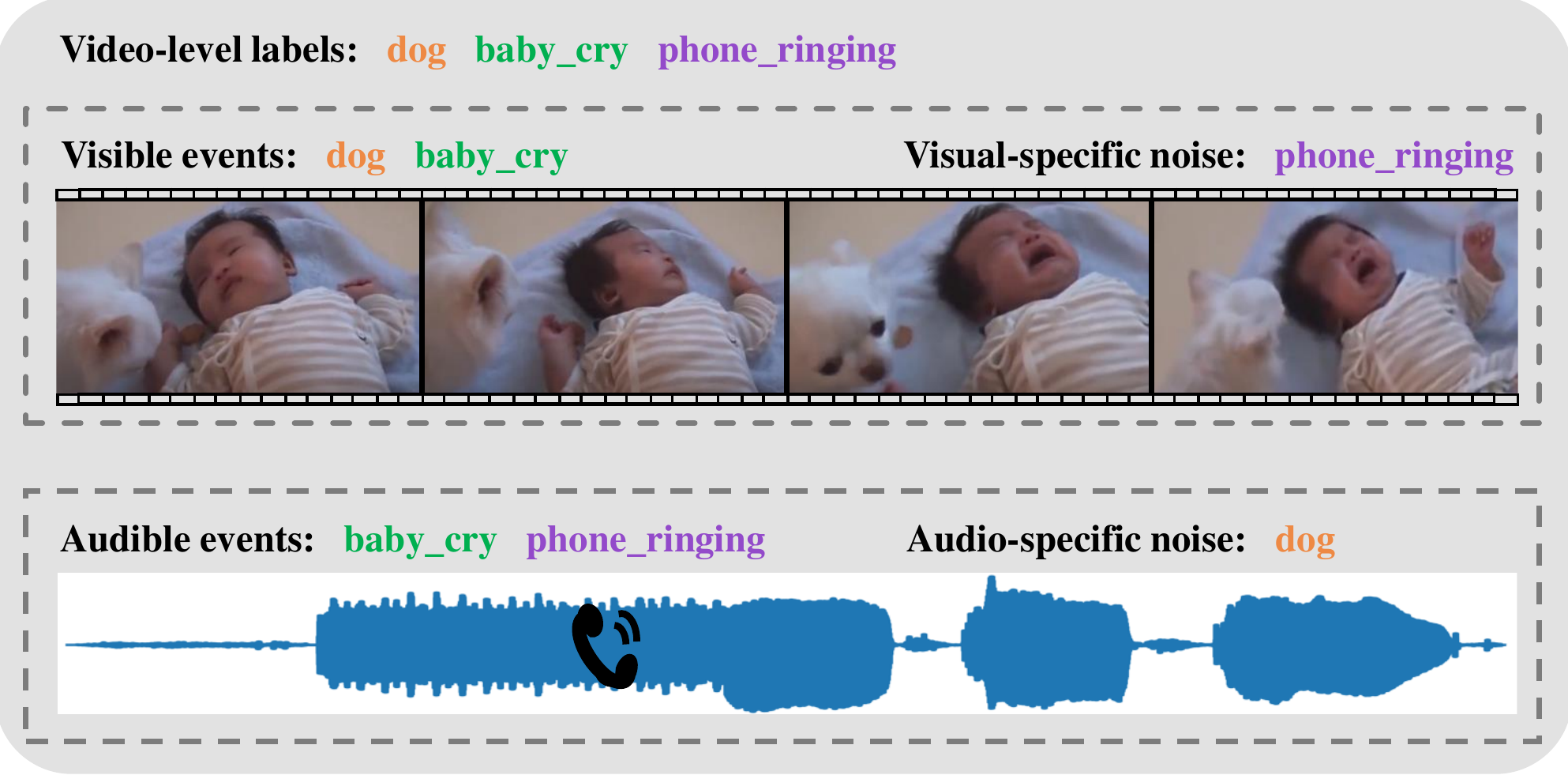}
  \caption{\textbf{An example to illustrate modality-specific noise in weakly-supervised AVVP task.} An infant accompanied by a dog is sleeping. Then, the phone bell rings and it frightens the baby to cry. In this example, the visible events in the whole video are ``\,dog\,'' and ``\,baby\_cry\_infant\_cry\,''. The audible events are ``\,telephone\_bell\_ringing\,'' and ``\,baby\_cry\_infant\_cry\,''. Thus, the ``\,dog\,'' can be treated as audio-specific noise, and ``\,telephone\_bell\_ringing\,'' is regarded as visual-specific noise.}
  \label{fig:intro}
\end{figure}

AVVP task is formulated in a weakly-supervised manner since precisely annotating labels would be expensive and time-consuming. Event labels are provided for each video in the training set, but the events' detailed modality and temporal location are unavailable. This manner is termed Multimodal Multiple Instance Learning (MMIL).
The weakly-supervised setting and audio-visual inconsistency lead to a serious issue: \textbf{modality-specific noise}.
Modality-specific noise is related to \emph{the event clues that do not appear in one of the two modalities}. 
Taking Figure~\ref{fig:intro} as an example, the event ``telephone\_bell\_ringing'' only appears in the audio track and can not be perceived from the visual track, which thus is an improper supervised signal for visual modality and can be regarded as a visual-specific noisy label.
We argue that lowering the interference of modality-specific noise can significantly advance the quality of audio-visual video parsing.

The pioneers in audio-visual video parsing have made sustained efforts to learn the spatio-temporal clues under such a weakly-supervised setting. HAN~\cite{tian2020unified} adopts a hybrid attention network optimized with a label smoothing mechanism, but it still suffers from the modality-specific noise.
MA~\cite{wu2021exploring} exchanges audio or visual tracks between two unrelated videos to yield reliable event labels for each modality.
However, since the cross-modal aggregation in \cite{wu2021exploring} is trained by paired audio-visual features, this might interfere with the uncertainty-assessing procedure on their re-assembled videos. 
As the previous label denoising methods~\cite{wei2020combating,yu2019does,nguyen2020self,Li2020DivideMixLW} focus on unimodal patterns, leveraging the cross-modal correlation to alleviate modality-specific noise still remains to be studied further. Consequently, we design a novel training strategy to mitigate the impacts of modality-specific noise for weakly-supervised audio-visual video parsing.

In this paper, we propose the \textbf{Joint-Modal Label Denoising (JoMoLD)} training strategy to dynamically alleviate \textbf{modality-specific noise} through careful loss analysis for both modalities.
We take the inspiration from two observations. 1) Neural networks tend to learn cleanly labeled samples first, and gradually memorize noisy ones~\cite{zhang2017understanding,arpit2017closer,li2020gradient}.
As a result, most noisily labeled data would be more challenging to learn than correctly labeled ones. 
From the view of loss patterns, the loss of cleanly labeled samples would be lower than noisily labeled ones.
2) Under the weakly supervised training setting, an event label ought not to serve as noise for both modalities, \textit{i.e.}, this event appears in at least one modality. 
According to these observations, the loss of a noisily labeled modality tends to be higher than the loss of the other modality where the event appears. 
We call this phenomenon \emph{loss inconsistency among different modalities}.

Based on the above analysis, we leverage audio and visual loss patterns to remove modality-specific noisy labels for each modality.
First, we design a noise estimator to approximately pre-estimate the noise ratio per category individually for each modality, which guides our proposed training strategy in determining the modality-specific noise. 
Then, when training the parsing model, we rank the losses within a mini-batch separately for two modalities.
Based on the pre-estimated noise ratios, we treat the labels with inconsistent losses between two modalities as modality-specific noise.
For each iteration in the training phase, we remove modality-specific noisy labels from their corresponding modality before back-propagation.
This training strategy makes the parsing network robust to modality-specific noise under the weakly-supervised setting of AVVP. 
Our experiment results significantly outperform the previous state of the arts, which validates the effectiveness of our proposed method.

In summary, we make the following contributions:
\begin{itemize}
    \item \emph{First}, we develop a noise ratio estimator to calculate the modality-specific noise ratios, which play a crucial role in determining the noise selection. 
    \item \emph{Second}, we propose a general and dynamic training paradigm, namely Joint-Modal Label Denoising (JoMoLD),  to alleviate the issue of modality-specific noise from the perspective of joint-modal label denoising on weakly-supervised AVVP task.
    \item \emph{Finally}, the experiments on the LLP dataset validate the effectiveness of our method. 
    Especially, the segment-level visual metric is improved from 60.0\% to 63.8\% over the state of the art. 
\end{itemize}

\section{Related Work}
\label{sec:related}
\subsection{Audio-Visual Learning}
Audio-visual joint learning has derived a variety of tasks, such as audio-visual representation learning~\cite{aytar2016soundnet,gupta2016cross,arandjelovic2017look,korbar2018cooperative,alwassel2020self}, audio-visual sound separation~\cite{gao2021visualvoice,ephrat2018looking,zhao2018sound,zhou2020sep}, sound source localization~\cite{senocak2018learning,arandjelovic2018objects}, audio-visual video captioning~\cite{rahman2019watch,hori2017attention,tian2018attempt}, audio-visual event localization~\cite{lin2019dual,wu2019dual,zhou2021positive}, and audio-visual action recognition~\cite{xiao2020audiovisual,panda2021adamml,gao2020listen}. Most of these works are based on the assumption that audio and visual signals are always semantically corresponding and temporally synchronized. 

For audio-visual representation learning,~\cite{aytar2016soundnet} and~\cite{gupta2016cross} transfer discriminative visual knowledge from pre-trained visual models into the audio modality.
Alwassel \etal~\cite{alwassel2020self} leverage unsupervised clustering results within one modality as a supervised signal to the other modality.
Other works explore learning instance-level audio-visual correspondence. Arandjelovic \etal~\cite{arandjelovic2017look} design a pretext task to learn correspondent representations between images and audio.
Korbar \etal~\cite{korbar2018cooperative} further utilize the temporal synchronization between audio and visual streams.

\subsection{Weakly-Supervised AVVP}
Audio-visual video parsing (AVVP) task~\cite{tian2020unified} breaks the restriction that audio and visual signals are definitely aligned. 
Due to the difficulty of exhaustive manual annotations, the audio-visual video parsing task is under the weakly-supervised setting, where only the video-level labels are provided for training. 

To tackle the weakly-supervised AVVP task, previous work~\cite{tian2020unified} proposes a hybrid attention network and attentive Multimodal Multiple Instance Learning (MMIL) Pooling mechanism to aggregate all features. 
Wu \etal~\cite{wu2021exploring} refine individual modality labels by swapping audio or visual tracks between two unrelated videos. Nevertheless, they independently refine modality labels without considering the relationship between the two modality labels.
Our work obtains more precise modality-specific labels in a joint-modal label denoising manner.

\subsection{Learning With Label Noise}
Deep neural networks have been demonstrated to learn clean samples first, and gradually memorize samples with noisy labels~\cite{zhang2017understanding,arpit2017closer,li2020gradient,han2018co}.
Recent works~\cite{liu2020early} show that such early-learning and memorization phenomena can even be observed in linear models. Many works are trying to handle this problem from different perspectives.
One kind of methods~\cite{zhang2018generalized,wang2019symmetric,kim2021joint} designs reasonable loss functions or regularization mechanisms to reduce overfitting noise. Semi-supervised learning is also adopted in some works~\cite{Li2020DivideMixLW,mandal2020novel}.

Two types of methods related to our approach are learning on selected clean samples~\cite{malach2017decoupling,han2018co,yu2019does,nguyen2020self} and correcting noisy labels~\cite{arazo2019unsupervised,tanaka2018joint,yi2019probabilistic,song2019selfie}.
Han \etal~\cite{han2018co} propose ``Co-teaching'', which optimizes two models on clean samples selected by the paired network.
To prevent the two networks converging to a consensus, ``Co-teaching+''~\cite{yu2019does} is proposed to combine ``Co-teaching'' with ``Update by Disagreement''~\cite{malach2017decoupling} strategy.
These collaborative learning with label noise methods are motivated by ``Co-training''~\cite{blum1998combining} in semi-supervised training.
Tanaka \etal~\cite{tanaka2018joint} utilize model output to reassign labels for noisy samples. However, most of these works focus on single-modal label denoising, and cannot be directly utilized for multi-modal label denoising.

Some multi-modal learning works~\cite{amrani2021noise,hu2021learning} also try to mitigate the effects of noisy labels.
Amrani \etal~\cite{amrani2021noise} propose to reduce multi-modal noise estimation to a multi-modal density estimation problem. Hu \etal~\cite{hu2021learning} propose a ``Robust Clustering loss'' to make the model focus on clean samples instead of noisy ones. However, these methods do not explore the correlation between multi-modalities and have not addressed the multi-modal noisy labels problem.

Our work deals with the issue of modality-specific noisy labels in weakly-supervised AVVP task. We aim to generate more reliable modality-specific labels considering cross-modal connections.

\section{Method}
Our main idea is to utilize the cross-modal loss patterns to perceive modality-specific noise, which is compatible with off-the-shelf networks (\eg, \cite{tian2020unified}) on weakly-supervised AVVP task. 
In this section, we elaborately introduce our proposed method, \ie, \textbf{Joint-Modal Label Denoising} (JoMoLD).
Firstly, we formulate the problem statement along with the baseline framework in Sec.~\ref{sec:prelim}.
Secondly, the noise estimator is proposed to calculate modality-specific noise ratios in Sec.~\ref{sec:estimate}.
Thirdly, we design an effective algorithm to remove modality-specific noisy labels by pre-estimated noise ratios in Sec.~\ref{sec:denoise}.
Lastly, we give an insightful discussion of our method in Sec.~\ref{sec:discussion}.

\subsection{Preliminaries}
\label{sec:prelim}
\noindent\textbf{Problem Statement.} 
This task aims to detect audible events or visible events that appear in each segment of a video, which is formulated as a Multimodal Multiple Instance Learning (MMIL) problem with $C$ event categories by Tian \etal~\cite{tian2020unified}.
Specifically, given a T-second video sequence $\{A_t, V_t\}_{t=1}^T$, $A_t$ denotes $t$-$th$ segment in audio track and $V_t$ denotes $t$-$th$ segment in visual track. During evaluation, let $(y_t^a, y_t^v, y_t^{av})$ denote audio, visual and audio-visual event labels at $t$-$th$ segment, respectively. Note that $y_t^a$, $y_t^v$ and $y_t^{av}$ are $\{0,1\}^C$ vectors, indicating the presence or absence of each event category. The audio-visual event represents this event occurs in both visual track and audio track at time $t$.
However, as a weakly-supervised task, we can only access video-level event label $y \in \{0,1\}^{C}$ instead of accurate segment-level labels during training. In other words, we only know which events occurred in a video, but can not acquire when events occur and in which modality the events appear.
Following the practice in \cite{tian2020unified}, we use pre-trained off-the-shelf networks to extract local audio and visual features $\{f_t^a, f_t^v\}_{t=1}^T$ for each segment.

\noindent\textbf{Baseline Framework.} 
We here use previous work~\cite{tian2020unified} (denoted as $\mathcal{F}$) as our important baseline. To capture temporal context and leverage the clues in different modalities, Tian~\etal~\cite{tian2020unified} adopt self-attention and cross-attention mechanisms to aggregate inner-modal and intra-modal information on segment features. Furthermore, an attentive MMIL Pooling mechanism is proposed to yield modality-level and video-level predictions.
As a multi-label multi-class learning task, it naturally uses binary cross-entropy (BCE) loss to optimize the model. In the baseline, the video-level label $y$ is used to supervise both modality-level and video-level predictions.
Due to the modality-specific noise, we argue that such a fashion is misleading for model training.
Therefore, we develop a dynamic training strategy, termed \textbf{Joint-Modal Label Denoising} (JoMoLD), to alleviate the effect of modality-specific noise during training.

\subsection{Estimating Noise Ratios}
\label{sec:estimate}
Our algorithm requires pre-estimating the modality-specific noise ratio per category, which assists our model in determining which labels should be removed during training. Since the real modality-specific noise ratios are unavailable in the training phase, we design a simple yet effective manner to approximately estimate noise ratios. The key insight is that the baseline model trained on this task already has a certain capacity for discriminating noisy labels. 

Specifically, we train a noise estimator $\mathcal{H}$ following baseline \cite{tian2020unified} with one notable modification: removing the cross-modal attention in the overall pipeline. As cross-modal attention exchanges the information between audio and visual features, it practically interferes with the noise estimation for each modality. Experiment in Table~\ref{tab:cross-modal attention} has proved this point.
Let $\hat{\bf{P}}^a,\hat{\bf{P}}^{v}\in \mathbb{R}^{N\times C}$ denote the audio and visual predictions of $\mathcal{H}$, and $\bar{{P}}^a,\bar{{P}}^v\in\mathbb{R}^C$ are the mean of predictions for each category.
${\bf{Y}} \in \{0,1\}^{N \times C}$ are video-level labels. Note that $N$ is the number of videos in training set.
Then, we need to define which labels are probably noise.
For example, we argue the annotated $c$-$th$ category label for $i$-$th$ video, \ie~${\bf{Y}}[i,c]=1$, is not reliable for audio modality if $\hat{\bf{P}}^a[i, c]/{\bar{P}}^a[c]$ is lower than a pre-set threshold $\theta^a$. 
Here, $\hat{\bf{P}}^a[i, c]$ is normalized by ${\bar{P}}^a[c]$ so as to alleviate the impact of imbalanced distribution of predictions in each event category.
The procedure of estimating noise ratio for $c$-$th$ category is summarized as follows:
\begin{equation}
\begin{aligned}
{{\bf{r}}^a[c] = \frac{\sum_{i=1}^{N} \mathbb{I}(\hat{\bf{P}}^a[i, c]/{\bar{P}}^a[c] < \theta^a) \times {\bf{Y}}[i, c]}{\sum_{i=1}^N{\bf{Y}}[i, c]}}, \\
{{\bf{r}}^v[c] = \frac{\sum_{i=1}^{N} \mathbb{I}(\hat{\bf{P}}^v[i,c]/{\bar{P}}^v[c] < \theta^v) \times {\bf{Y}}[i, c]}{\sum_{i=1}^N{\bf{Y}}[i, c]}},\\
\end{aligned}
\end{equation}
where ${\bf{r}}^a \,({\bf{r}}^v) \in \mathbb{R}^{C}$ denotes noise ratios of positive labels in audio (visual) track for $C$ categories,
$\mathbb{I}$ is the indicator function, and ``$\times$'' denotes multiplication. Note that $\theta^a$ ($\theta^v$) can be seen as the confidence to determine noisy labels for audio (visual) modality. Table~\ref{tab:thres} shows that final performance of our proposed method is not sensitive to $\theta^a$ and $\theta^v$. 
The estimated noise ratios will be used as a priori knowledge for the label denoising procedure. 

Generally, our noise estimator is essentially different from Wu \etal ~\cite{wu2021exploring}. 
They exchange the audio tracks between two unrelated videos to filter out the noisy labels and re-train the model from scratch based on refined labels.
On the one hand, as cross-modal attention aggregates multi-modal clues, the modality-level predictions would affect each other. Thus it is improper to assess the uncertainty of event labels for each modality when using cross-modal attention. 
On the other hand, the refined labels will be fixed after the label refinement procedure in \cite{wu2021exploring}. It causes that even the wrongly refined labels would also be used to re-train the parsing model in the whole re-training phase.
In contrast, our method alleviates the potential negative impacts of cross-modal attention in the phase of noise estimation. 
Furthermore, modality-specific noisy labels are removed dynamically, which is expected to tolerate some improper refinements in the previous training.
Even though the intrinsic biases might exist in the noise estimator, the estimated noise ratios are still effective for our modality-specific label denoising algorithm, which is also validated in experiments.

\begin{figure}[t]
  \centering
  \includegraphics[width=12cm,height=6cm]{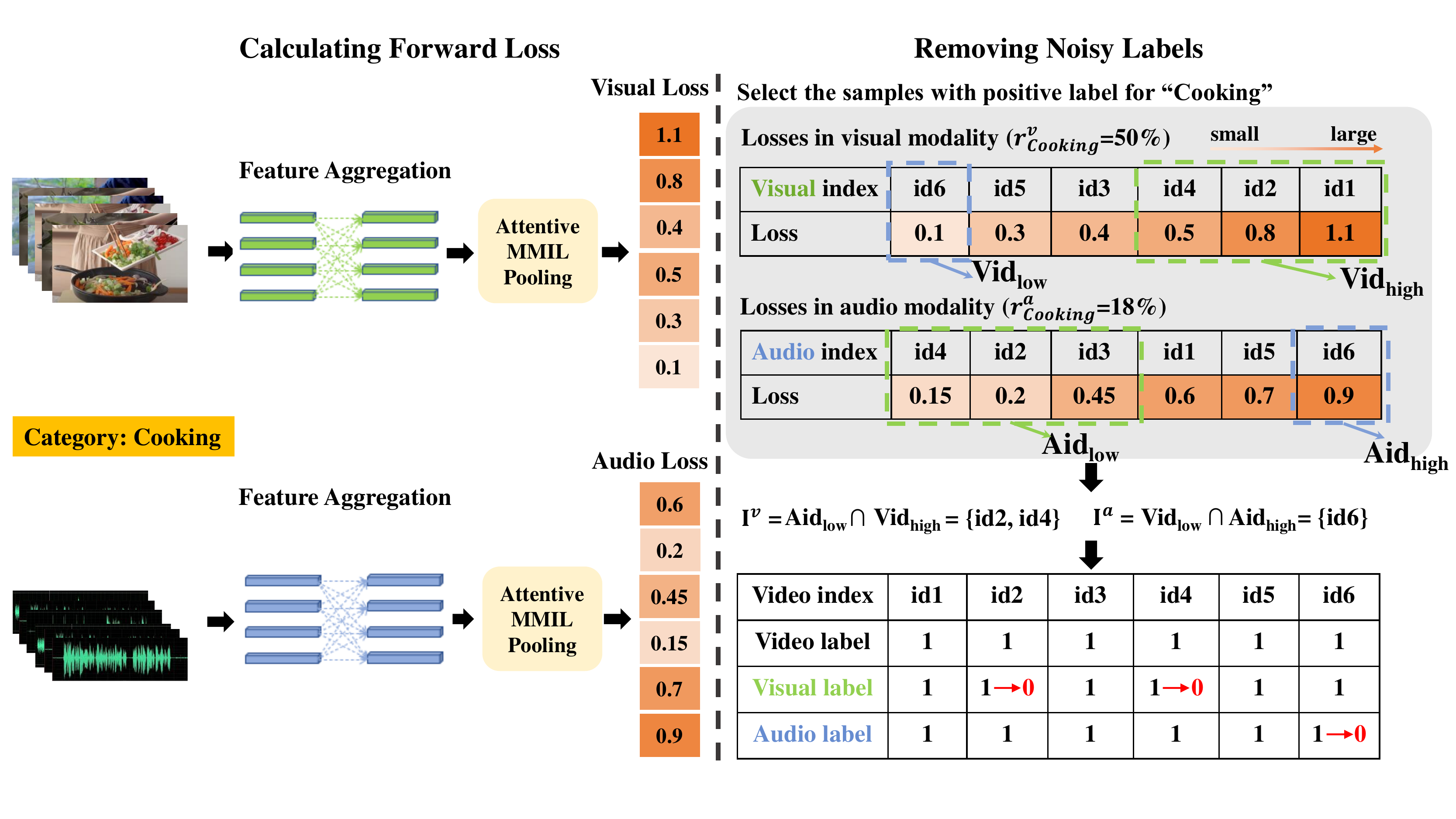}
  \caption{\textbf{The proposed modality-specific label denoising procedure.} 
  The label denoising procedure consists of \textbf{Calculating Foward Loss} and \textbf{Removing Noisy Labels}. In this case, we represent the label denoising process for the ``Cooking'' event in a batch of videos.
  In calculating forward loss, we aggregate the intra-modal features, obtain the modality-level predictions, and further calculate the modality losses.
  Based on the estimated noise ratios ${\bf{r}}^v_{Cooking}$ and the sorted visual losses, we obtain the indices of noisy samples for visual modality, \ie\, ${\bf{I}}^v$.
  Then we remove the video labels for videos by ${\bf{I}}^v$ to generate refined visual labels. In the figure, ``\,${\color{red}\rightarrow0}$\,'' denotes the label of ``Cooking'' is removed.
  The same procedure is applied to audio modality label denoising.}
  \label{fig:method}
\end{figure}

\subsection{Modality-specific Label Denoising}
\label{sec:denoise}

In this section, our goal is to remove modality-specific noisy labels and optimize the parsing model $\mathcal{F}$ with the refined labels. Therefore, we need to solve a tricky issue: how to identify the modality-specific noisy labels.
As illustrated in Figure~\ref{fig:method}, our modality-specific label denoising procedure can be summarized as two steps: a) \textbf{Calculating Forward Loss} and b) \textbf{Removing Noisy Labels}. 
As a general training strategy, we adopt the model in~\cite{tian2020unified} as our backbone network $\mathcal{F}$ to verify our method. 

First, in the step of \textbf{Calculating Forward Loss}, losses are calculated in each modality for removing the noisy labels.
Specifically, we feed the extracted local features of a batch of videos into the network, and calculate the BCE losses for each modality. Here, cross-modal attention is skipped in this step to avoid interference of cross-modal feature aggregation. Note that the losses calculated in this step are only utilized to remove noisy labels but not optimize the network.
Let $B$ represent the batch size, ${\bf{L}}^a, {\bf{L}}^v \in \mathbb{R}^{B \times C}$ denote the BCE losses in audio and visual modalities.

\definecolor{dg}{rgb}{0.0, 0.5, 0.0}
\begin{algorithm}[t]
\caption{The Pipeline of JoMoLD}
\label{algo:denoise}
\begin{algorithmic}[1]
{\fontsize{8pt}{8pt}{
\REQUIRE ~~\\
Noise ratios ${\bf{r}}^a \in \mathbb{R}^C,{\bf{r}}^v\in \mathbb{R}^C$ estimated in Sec.~\ref{sec:estimate} \\
Total training iterations $\Gamma$, the number of categories $C$, the batch size $B$ \\
The parsing network $\mathcal{F}$
\FOR{$i=0$ to $\Gamma$ - 1}
\STATE Fetch a mini-batch $\mathcal{B}$, and video-level labels ${\bf{Y}} \in \{0,1\}^{B \times C}$, ${\bf{Y}}^a$ = ${\bf{Y}}$, ${\bf{Y}}^v$ = ${\bf{Y}}$
\STATE Feed $\mathcal{B}$ into $\mathcal{F}$ (skipping cross-modal attention) to calculate forward loss ${\bf{L}}^{a}$, ${\bf{L}}^{v} \in \mathbb{R}^{B \times C}$
    \FOR{$c=1$ to $C$}
    \STATE\label{select}
    \textcolor{dg}{\# Find the indexes of positive labels and the number of positive samples} \\
    ${\bf{I}} = nonzero({\bf{Y}}[:, c]),\quad B' = \sum_{i=1}^B{\bf{Y}}[:, c]$ \\
    \textcolor{dg}{\# Calculate the numbers of candidate noise for audio and visual modalities} \\
    ${\bf{M}}^a=int({\bf{r}}^a[c]\times B'),\quad {\bf{M}}^v=int({\bf{r}}^v[c]\times B')$ \\
    \textcolor{dg}{\# Determine the indexes of audio noise in batch $\mathcal{B}$} \\
    ${\bf{I}}^a = {\bf{I}}[\mathcal{G}({\bf{-L}}^a[{\bf{I}}, c], {\bf{M}}^a)] \cap {\bf{I}}[\mathcal{G}({\bf{L}}^v[{\bf{I}}, c], {\bf{M}}^a)]$ \\
    \textcolor{dg}{\# Determine the indexes of visual noise in batch $\mathcal{B}$}\\
    ${\bf{I}}^v = {\bf{I}}[\mathcal{G}({\bf{-L}}^v[{\bf{I}}, c], {\bf{M}}^v)] \cap {\bf{I}}[\mathcal{G}({\bf{L}}^a[{\bf{I}}, c], {\bf{M}}^v)]$\\
    \STATE
    \textcolor{dg}{\# Remove noisy labels} \\
    ${\bf{Y}}^a[{\bf{I}}^a, c]=0$,\quad
    ${\bf{Y}}^v[{\bf{I}}^v, c]=0$
    \ENDFOR
\STATE Feed $\mathcal{B}$ into $\mathcal{F}$, and utilize ${\bf{Y}}$, ${\bf{Y}}^a$ and ${\bf{Y}}^v$ to optimize network $\mathcal{F}$
\ENDFOR
}}
\end{algorithmic}
\end{algorithm}

Second, in the step of \textbf{Removing Noisy Labels}, modality-specific noisy labels are determined according to the loss patterns based on estimated noise ratios ${\bf{r}}^a$ and ${\bf{r}}^v$. 
We here define a function for better introducing the procedure of removing labels: 

\begin{equation}
\begin{aligned}
\mathcal{G}(\mathcal{L}, n) &=
argsort(\mathcal{L})[0:n] ,\\
\end{aligned}
\end{equation}
where $\mathcal{L}$ denotes the losses of a batch of samples, and $n$ is the parameter to control the number of selected indexes. $argsort(\mathcal{L})$ is a function that sorts $\mathcal{L}$ in ascending order and returns the indexes of sorted losses. We use $argsort(-\mathcal{L})$ to obtain the indexes of losses $\mathcal{L}$ sorted in descending order. The $nonzero(\cdot)$ is a function that returns the indexes of samples whose value is not 0. The procedure to determine which labels are modality-specific noise corresponds to Step~\ref{select} in Algorithm~\ref{algo:denoise}. Intuitively, taking audio-specific label denoising as an example, we argue that a positive label with 
a high loss in the audio modality and a low loss in visual modality would be an audio-specific noisy label. 
We have discussed the interpretability and reasonability of the way to determine noisy labels in Sec.~\ref{sec:discussion}.
The identified modality-specific noisy labels are removed from ${\bf{Y}}$, to generate refined modality labels ${\bf{Y}}^a$ and ${\bf{Y}}^v$. 

Lastly, in each training iteration, we feed the batch $\mathcal{B}$ to network $\mathcal{F}$ to obtain predictions. The refined labels ${\bf{Y}}^a$, ${\bf{Y}}^v$ and ${\bf{Y}}$ serve as supervised signals for audio predictions, visual predictions, and video predictions, respectively. Experimental results in Sec.~\ref{sec:exp} have validated the effectiveness of our method.

\subsection{Discussion}
\label{sec:discussion}
As we have mentioned above, we first train the noise estimator $\mathcal{H}$ to estimate the modality-specific noise ratios for each category. $\mathcal{H}$ is modified from the baseline $\mathcal{F}$~\cite{tian2020unified} by removing cross-modal attention. After that, we adopt the noise ratios to guide the training process. We use the network $\mathcal{F}$ as our parsing model, but skip the cross-modal attention 
during \textbf{calculating forward loss}. Finally, there is still confronted with two critical questions about the motivation of our method:

\emph{1) Why do we regard the event labels with higher losses as the candidate noise set by using intra-modal loss patterns}?
As analyzed previously, deep neural networks are prone to learn clean labels first, but over-fit the noisy labels with more training epochs~\cite{arpit2017closer,zhang2017understanding,li2020gradient}. In other words, the losses of clean labels would drop faster than noisy labels. Built upon this observation, we argue that noisy labels are more likely to exist in the samples with higher losses.

\emph{2) Why do cross-modal cues make more evident improvement for determining the modality-specific noise?}
If we directly treat the event labels with higher losses as noise, some hard labels would not be seen by the model during training in extreme circumstances. 
We observe that a video-level event label typically appears in at least one modality.
In this sense, if one label has a higher loss in audio modality but a lower loss in the visual modality, it means evident clues have appeared in visual rather than audio modality. Therefore,  
we speculate that this label may be noisy for audio modality with high confidence. 
It is effective to utilize the complementary knowledge to recheck label noise, which is also verified in Table~\ref{tab:self-modal}.

\section{Experiments}
\label{sec:exp}
This section elaborates on our experiments' details and compares our proposed JoMoLD with state-of-the-art methods. In the ablation studies, we present the effect of each module. We conduct a qualitative analysis and show the advantages of our JoMoLD over state-of-the-art methods.

\subsection{Experiment Settings}
\noindent\textbf{Dataset.} We evaluate our method for weakly-supervised AVVP task on the $\textit{Look, Listen, and Parse}$ (LLP) dataset. It consists of 11849 10-second videos with 25 event categories. The categories cover a wide range of domains such as human activities, animal activities, music performance, \etc~
We utilize the official data split for training and evaluation. There are 10000 videos for training and 1849 validation-test videos for evaluation.

\noindent\textbf{Evaluation Metrics.} We evaluate the parsing performance of all events (audio, visual, and audio-visual events) under segment-level and event-level metrics. 
We use both segment-level and event-level F-scores as metrics. The former metrics evaluate the segment-wise prediction performance. The latter metrics are designed to extract events with consecutive positive snippets in the same event categories, and 0.5 is used as the mIoU threshold to compute event-level F-scores.
Moreover, we also assess the comprehensive performance for all events by ``Type@AV'' and ``Event@AV'' metrics. Type@AV is calculated by averaging audio, visual, and audio-visual metrics. Event@AV computes the results considering all audio and visual events instead of averaging metrics. 
Abbreviations of metric names are represented in all experiment tables, where ``A'' denotes audio events, ``V'' represents visual events, ``AV'' denotes audio-visual events, ``Type'' indicates Type@AV, and ``Event'' denotes ``Event@AV''.

\noindent\textbf{Implementation Details.}
Following the data preprocessing in \cite{tian2020unified,wu2021exploring}, 
we decode a 10-second video into 10 segments, and each segment contains 8 frames.
We use pre-trained ResNet152~\cite{he2016deep} and R(2+1)D~\cite{tran2018closer} to capture the appearance and motion features and concatenate them as low-level visual features. For audio, we adopt pre-trained VGGish~\cite{hershey2017cnn} to yield audio features. Adam optimizer is used to train the model, and the learning rate 5e-4 drops by a factor of 0.25 for every 6 epochs. We train the model for 25 epochs with batch size 128.

\begin{table*}[tb!]
\centering
\caption{\textbf{Comparisons with the state-of-the-art methods on the test set of LLP.} JoMoLD achieves the best performance among them. 
``CL'' denotes the contrastive learning proposed in MA~\cite{wu2021exploring}.
We simply add ``CL'' loss into the existing loss functions when optimizing the network, but do not utilize it in label denoising.
Results of our method combined with ``CL'' proves the flexibility and effectiveness of JoMoLD.
``-'' denotes this result is not available.}
\vspace{3mm}
\resizebox{\linewidth}{!}{
    \begin{tabular}{cl|ccccc|ccccc}
    \toprule
    \multicolumn{2}{c|}{} & \multicolumn{5}{c|}{Segment-level} & \multicolumn{5}{c}{Event-Level} \\
    \multicolumn{2}{c|}{\multirow{-2}{*}{Methods}} & A & V & AV & Type & Event & A & V & AV & Type & Event \\ \hline
    \multicolumn{2}{c|}{TALNet~\cite{wang2019comparison}} & 50.0 & - & - & - & - & 41.7 & - & - & - & - \\
    \multicolumn{2}{c|}{STPN~\cite{nguyen2018weakly}} & - & 46.5 & - & - & - & - & 41.5 & - & - & - \\
    \multicolumn{2}{c|}{CMCS~\cite{liu2019completeness}} & - & 48.1 & - & - & - & - & 45.1 & - & - & - \\ \cline{1-12}
    \multicolumn{2}{c|}{AVE~\cite{tian2018audio}} & 49.9 & 37.3 & 37.0 & 41.4 & 43.6 & 43.6 & 32.4 & 32.6 & 36.2 & 37.4 \\
    \multicolumn{2}{c|}{AVSDN~\cite{lin2019dual}} & 47.8 & 52.0 & 37.1 & 45.7 & 50.8 & 34.1 & 46.3 & 26.5 & 35.6 & 37.7 \\
    \multicolumn{2}{c|}{HAN~\cite{tian2020unified}} & 60.1 & 52.9 & 48.9 & 54.0 & 55.4 & 51.3 & 48.9 & 43.0 & 47.7 & 48.0 \\ \cline{1-12}
    
    \multicolumn{2}{c|}{HAN \textit{w/} Co-teaching+ (HC)~\cite{yu2019does}} & 59.4 & 56.7 & 52.0 & 56.0 & 56.3 & 50.7 & 53.9 & 46.6 & 50.4 & 48.7 \\
    
    \multicolumn{2}{c|}{HAN \textit{w/} JoCoR (HJ)~\cite{wei2020combating}} & 61.0 & 58.2 & 53.1 & 57.4 & 57.7 & 52.8 & 54.7 & 46.7 & 51.4 & 50.3 \\ \cline{1-12}
    
    \multicolumn{1}{c|}{} & \textit{w/o}~CL & 59.8 & 57.5 & 52.6 & 56.6 & 56.6 & 52.1 & 54.4 & 45.8 & 50.8 & 49.4 \\
    \multicolumn{1}{c|}{\multirow{-2}{*}{MA~\cite{wu2021exploring}}} & \textit{w/}~CL & 60.3 & 60.0 & 55.1 & 58.9 & 57.9 & 53.6 & 56.4 & 49.0 & 53.0 & 50.6 \\ \cline{1-12}
    
    \multicolumn{1}{c|}{} & \textit{w/o}~CL & \begin{tabular}[c]{@{}c@{}}60.6\end{tabular} & \begin{tabular}[c]{@{}c@{}}62.2\end{tabular} & \begin{tabular}[c]{@{}c@{}}56.0\end{tabular} & \begin{tabular}[c]{@{}c@{}}59.6\end{tabular} & \begin{tabular}[c]{@{}c@{}}58.6\end{tabular} & \begin{tabular}[c]{@{}c@{}}53.1\end{tabular} & \begin{tabular}[c]{@{}c@{}}58.9\end{tabular} & \begin{tabular}[c]{@{}c@{}}49.4\end{tabular} & \begin{tabular}[c]{@{}c@{}}53.8\end{tabular} & \begin{tabular}[c]{@{}c@{}}51.4\end{tabular} \\
    \multicolumn{1}{c|}{\multirow{-2}{*}{\textbf{JoMoLD (Ours)}}} & \textbf{\textit{w/}~CL} & \begin{tabular}[c]{@{}c@{}}\textbf{61.3}\end{tabular} & \begin{tabular}[c]{@{}c@{}}\textbf{63.8}\end{tabular} & \begin{tabular}[c]{@{}c@{}}\textbf{57.2}\end{tabular} & \begin{tabular}[c]{@{}c@{}}\textbf{60.8}\end{tabular} & \begin{tabular}[c]{@{}c@{}}\textbf{59.9}\end{tabular} & \begin{tabular}[c]{@{}c@{}}\textbf{53.9}\end{tabular} & \begin{tabular}[c]{@{}c@{}}\textbf{59.9}\end{tabular} & \begin{tabular}[c]{@{}c@{}}\textbf{49.6}\end{tabular} & \begin{tabular}[c]{@{}c@{}}\textbf{54.5}\end{tabular} & \begin{tabular}[c]{@{}c@{}}\textbf{52.5}\end{tabular} \\
    \bottomrule
    \end{tabular}
}
\label{tab:sota}
\end{table*}

\subsection{Comparison with State-of-the-art Methods}

We compare our method with different types of methods: weakly-supervised sound event detection methods TALNet~\cite{wang2019comparison}, weakly-supervised action localization methods STPN~\cite{nguyen2018weakly} and CMCS~\cite{liu2019completeness}, modified audio-visual event localization methods AVE~\cite{tian2018audio} and AVSD~\cite{lin2019dual}, the state-of-the-art AVVP methods HAN~\cite{tian2020unified} and MA~\cite{wu2021exploring}. 
In addition, Co-teaching+~\cite{yu2019does} and JoCoR~\cite{wei2020combating} are famous for learning with label noise methods that focus on single-modal tasks but not multi-modal tasks.
On this weakly-supervised AVVP task, we reproduce the variants of these two methods~\cite{yu2019does,wei2020combating} utilizing the backbone in HAN to compare with our method. 
The variants of these two methods are denoted as ``\,HAN \textit{w/} Co-teaching+\,'' (\textit{abbr}. HC) and ``\,HAN \textit{w/} JoCoR\,'' (\textit{abbr}. HJ).

Table~\ref{tab:sota} shows the results of JoMoLD and other state-of-the-art methods on the LLP test dataset.
Our JoMoLD here adopts the optimal settings studied by Sec.~\ref{sec:ablation}.
As a label denoising strategy, JoMoLD can combine with other feature learning methods to achieve higher performance, such as contrastive learning proposed in MA~\cite{wu2021exploring}.
Notably, our method outperforms the state-of-the-art methods (\eg, HC, HJ, and MA) by a non-negligible margin. For example, JoMoLD is higher than MA by 3.8 points in the segment-level visual event parsing metric.
These results demonstrate the effectiveness of our strategy of joint-modal label denoising.

\subsection{Ablation Studies}
\label{sec:ablation}
This section performs ablation studies on estimating noise ratios and modality-specific label denoising, respectively. Segment-level metrics are reported if not stated.
The optimal settings are explored by the following ablations.

\begin{table*}[h]
\centering
\caption{\small \textbf{Ablation studies.} Tables \ref{tab:noise_ratios}, \ref{tab:thres}, and \ref{tab:cross-modal attention} are studies on estimating noise ratios. Tables \ref{tab:self-modal}, \ref{tab:modality-only} and \ref{tab:warm-up} are ablations on modality-specific label denoising.}
    \begin{subtable}[thbp]{0.45\textwidth}
        \centering
        \caption{\textbf{Study the effectiveness of noise ratio estimator}. The constant noise ratios are hand-crafted.}
        \scalebox{0.8}{
        \begin{tabular}{c|ccccc}
                \toprule
                Noise Ratio & A & V & AV & Type & Event \\ \hline
                0.1 & 60.9 & 53.9 & 51.4 & 55.4 & 55.4 \\ 
                0.2 & 61.3 & 54.4 & 52.0 & 55.9 & 55.9 \\ 
                0.3 & 60.8 & 55.2 & 51.6 & 55.9 & 56.1 \\ 
                0.4 & 60.2 & 56.6 & 53.0 & 56.6 & 56.2 \\ 
                0.5 & 58.4 & 58.3 & 53.2 & 56.6 & 55.9 \\
                \textbf{Estimated ratios} & \textbf{61.3} & \textbf{63.8} & \textbf{57.2} & \textbf{60.8} & \textbf{59.9} \\
                \bottomrule
            \end{tabular}}
        \label{tab:noise_ratios}
    \end{subtable}
    \quad \quad
    \begin{subtable}[thbp]{0.45\textwidth}
        \centering
        \caption{\small \textbf{Study thresholds for noise estimation.} Segment-level Type@AV results are reported.}
        \scalebox{0.9}{
        \begin{tabular}{c|ccccc}
                \toprule
                \diagbox{audio $\theta^a$}{visual $\theta^v$} & 1.6 & 1.7 & 1.8 & 1.9 & 2.0\\
                \hline
                0.5 & 58.3 & 58.9 & 60.6 & 60.5 & 60.0\\
                0.6 & 58.9 & 58.8 & $\textbf{60.8}$ & 60.5 & 60.5\\
                0.7 & 58.5 & 59.0 & 60.8 & 60.7 & 60.4\\
                0.8 & 58.3 & 59.3 & 60.7 & 60.6 & 60.4\\
                \bottomrule
            \end{tabular}}
        \label{tab:thres}
    \end{subtable}
    \quad \quad \quad
    \begin{subtable}[thbp]{0.45\textwidth}
        \centering
        \caption{\small \textbf{Study the impact of cross-modal attention on noise estimator}. ``cm attn.'' represents cross-modal attention.}
    \scalebox{0.85}{
        \begin{tabular}{c|ccccc}
                \toprule
                Estimator & A & V & AV & Type & Event \\ \hline
                \textit{w/} cm attn. & 60.9 & 55.9 & 52.9 & 56.6 & 56.3 \\ 
                \textbf{\textit{w/o} cm attn.} & \textbf{61.3} & \textbf{63.8} & \textbf{57.2} & \textbf{60.8} & \textbf{59.9} \\ 
                \bottomrule
            \end{tabular}}
        \label{tab:cross-modal attention}
    \end{subtable}
    \quad \quad 
    \begin{subtable}[thbp]{0.45\textwidth}
        \centering
        \caption{\textbf{Intra-modal label denoising \textit{vs.} Joint-modal label denoising}. ``InMoLD'' indicates Intra-modal label denoising.}
        \scalebox{0.85}{
        \begin{tabular}{c|ccccc}
                \toprule
                Methods & A & V & AV & Type & Event \\ \hline
                InMoLD & 59.6 & 58.5 & 51.8 & 56.6 & 57.8 \\ 
                \textbf{JoMoLD} & \textbf{61.3} & \textbf{63.8} & \textbf{57.2} & \textbf{60.8} & \textbf{59.9} \\ 
                \bottomrule
            \end{tabular}}
        \label{tab:self-modal}
    \end{subtable}
    \quad \quad 
    \begin{subtable}[thbp]{0.45\textwidth}
        \centering
        \caption{\textbf{Single-modal label denoising \textit{vs.} joint-modal label denoising.} ``Audio only'' or ``Visual only'' denotes that label denoising is performed only for audio or visual track.}
        \scalebox{0.85}{
    \begin{tabular}{c|ccccc}
                \toprule
                Modality & A & V & AV & Type & Event \\ \hline
                Audio only & 61.3 & 53.2 & 50.3 & 54.9 & 56.2 \\
                Visual only & 61.0 & 62.7 & 56.2 & 60.0 & 59.2 \\
                \textbf{both (JoMoLD)} & \textbf{61.3} & \textbf{63.8} & \textbf{57.2} & \textbf{60.8} & \textbf{59.9} \\ \bottomrule
            \end{tabular}}
        \label{tab:modality-only}
    \end{subtable}
    \quad \quad \quad
    \begin{subtable}[thbp]{0.45\textwidth}
        \centering
        \caption{\small \textbf{Study the warm-up of noise ratios.} The noise ratios will be increased from 0 to its real values during the period of warm-up.}
        \scalebox{0.79}{
            \begin{tabular}{c|ccccc}
                \toprule
                Warm-up epochs& A & V & AV & Type & Event \\ \hline
                no warm-up & 60.6 & 63.3 & 56.3 & 60.1 & 59.0 \\
                0.7 & 61.1 & 63.6 & 56.8 & 60.5 & 59.7 \\
                0.8 & 60.9 & 63.8 & 56.6 & 60.4 & 59.5 \\
                \textbf{0.9} & \textbf{61.3} & \textbf{63.8} & \textbf{57.2} & \textbf{60.8} & \textbf{59.9} \\
                1.0 & 61.3 & 63.7 & 56.8 & 60.6 & 59.7 \\
                \bottomrule
            \end{tabular}}
        \label{tab:warm-up}
    \end{subtable}
    \label{tab:ablation_studies}
\end{table*}

\noindent\textbf{Ablation Studies on Estimating Noise Ratios: }

\textit{Study the Effectiveness of Noise Estimator.} To verify the importance of the noise estimator, we compare the performance of using a series of hand-crafted noise ratios to guide the label denoising procedure.
These manually set noise ratios contain no prior information.
As shown in Table~\ref{tab:noise_ratios}, 
our noise estimator provides sound guidance in determining noisy labels.

\textit{Study Thresholds for Noise Estimation.} 
We study the impact of thresholds, \ie, $\theta^a$ and $\theta^v$, in noise ratio estimation.
A lower threshold leads to smaller noise ratios and vice versa. Since the predictions of the noise estimator are normalized by the mean value per category, $\theta^a$ ($\theta^v$) may be more than 1. We find in Table~\ref{tab:thres} that our method is robust when $\theta^a$ and $\theta^v$ are within a reasonable range. 

\textit{Study the Impact of Cross-modal Attention on Noise Estimator.}
As shown in Table~\ref{tab:cross-modal attention},
it leads to a noticeable performance drop when training the noise estimator with cross-modal attention.
Since cross-attention mixes the clues between modalities, it causes improper noise estimation for each modality.
Experiments also verify the rationality of removing cross-attention in the noise estimator.

\noindent\textbf{Ablation Studies on Modality-Specific Label Denoising:}

\textit{Intra-modal Label Denoising \textit{vs.} Joint-modal Label Denoising.} 
Experiments in Table~\ref{tab:self-modal} demonstrates the superiority of JoMoLD over intra-modal label denoising (InMoLD).
The latter does not consider cross-modal clues for label denoising.
Specifically, for audio modality, InMoLD only considers the labels of samples with high losses in audio modality as audio noise, and does not check whether the losses in the visual modality of these samples are low. The procedure is the same for visual modality label denoising.
JoMoLD makes good use of the intuition that a label would not serve as noise for both modalities, so a confident noisy label should enjoy the different loss patterns between two modalities.

\textit{Single-modal Label Denoising \textit{vs.} Joint-modal Label Denoising.}
Single-modal label denoising adopts the same method as joint-modal label denoising, but conducts label denoising only for a single modality.
Table~\ref{tab:modality-only} shows that denoising labels for the visual track brings more improvement than the audio track.
This might be because LLP is an audio-dominant dataset with more visual noise. Moreover, removing noisy labels for both modalities brings further improvement.

\textit{Study Warm-up Strategy.} At the beginning of the training, the model treats clean and noisy labels alike, so we cannot distinguish them from the losses. In order to avoid selection bias in the early training, we adopt a warm-up strategy. It gradually increases the noise ratios from zero to the pre-computed values during warm-up epochs. The results are shown in Table~\ref{tab:warm-up}. Because warm-up strategy is not a critical step in JoMoLD, we omit it in Algorithm~\ref{algo:denoise} for clarity.

\begin{figure*}[t]
  \centering
  \includegraphics[width=12cm,height=5cm]{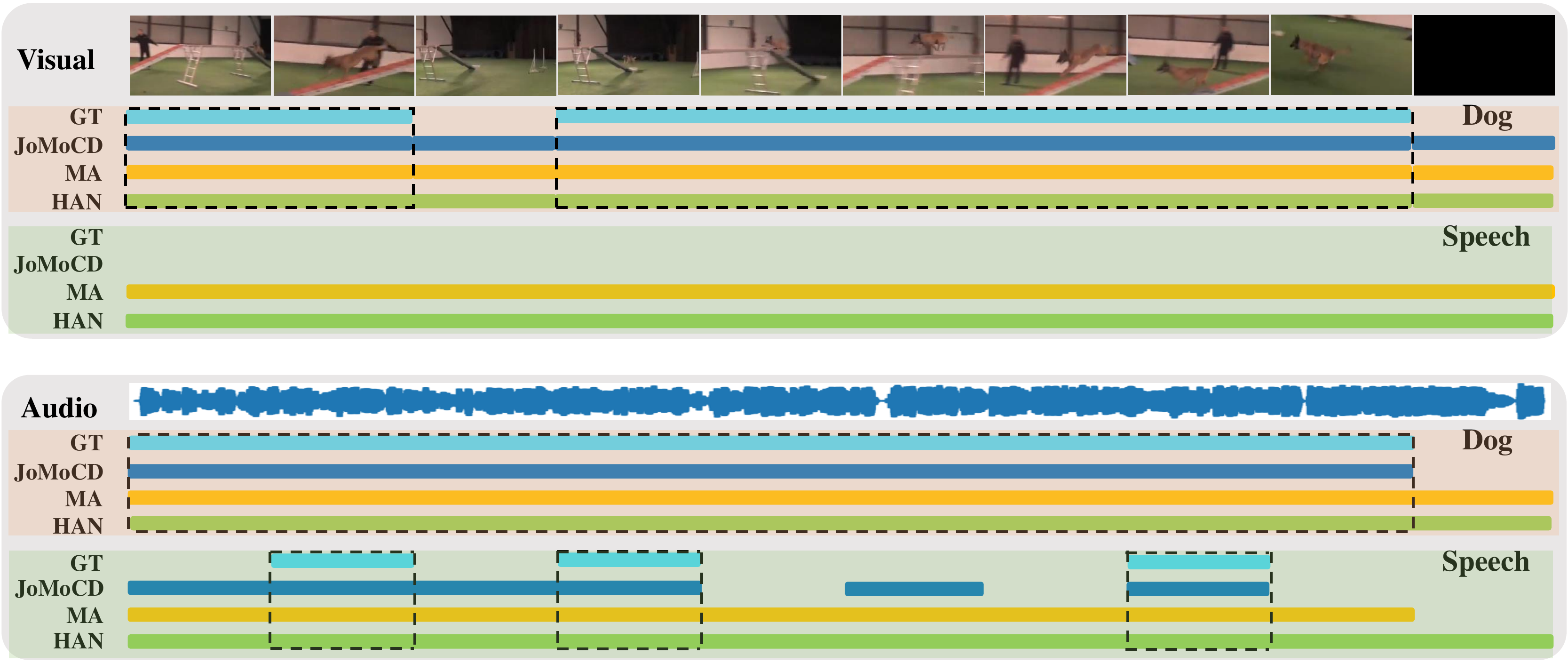}
  \caption{\textbf{Qualitative comparisons with the state-of-the-art mothods}. We detail the parsing visualization results of ``Dog'' and ``Speech'' categories for visual and audio modalities. ``GT'' denotes the ground truth annotations. We compare JoMoLD, MA~\cite{wu2021exploring} and HAN~\cite{tian2020unified} with GT. Generally, JoMoLD achieves better parsing results.}
  \label{fig:compare_parsing}
\end{figure*}

\subsection{Qualitative Analysis}

\noindent\textbf{Video Parsing Visualization.} Figure~\ref{fig:compare_parsing} visualizes the parsing results of JoMoLD, MA~\cite{wu2021exploring} and HAN~\cite{tian2020unified} as well as the ground truth annotation ``GT''. This video contains audio events ``Dog'' and ``Speech'', and visual event ``Dog''. Our method achieves the best performance in event recognition and localization among three methods.
In detail, MA and HAN wrongly predict the ``Speech'' event in the visual track while JoMoLD correctly predicts no ``Speech'' event in the visual track. This can be credited to our more accurate label denoising procedure during training. For the audio track, JoMoLD makes more precise detection results than the other two for the event ``Dog'' and ``Speech''.
Nevertheless, our method still makes mistakes for some segments. 
Due to the lack of segment-level supervised signals, there is still room for our method to improve performance.

\begin{figure}[t]
    \begin{subfigure}{.48\textwidth}
        \centering
        \includegraphics[width=\linewidth,height=3cm]{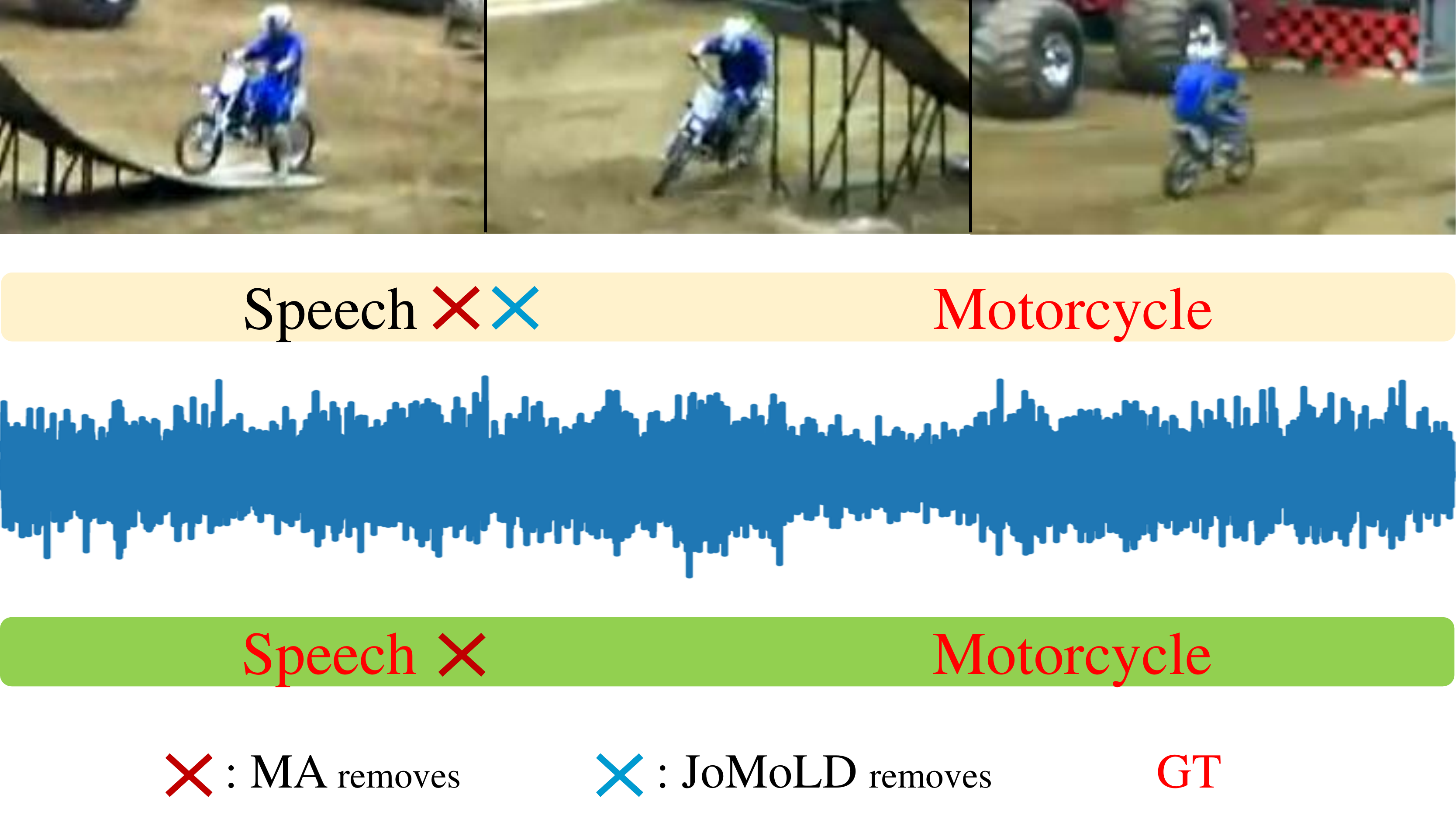}
        \caption{JoMoLD correctly removes the label ``Speech'' for visual track and remains it for audio track. But MA fails to recognize audio clues of ``Speech'' and removes it.}
        \label{fig:compare1}
    \end{subfigure}
    \hfill
    \begin{subfigure}{.48\textwidth}
        \centering
        \includegraphics[width=\linewidth,height=3cm]{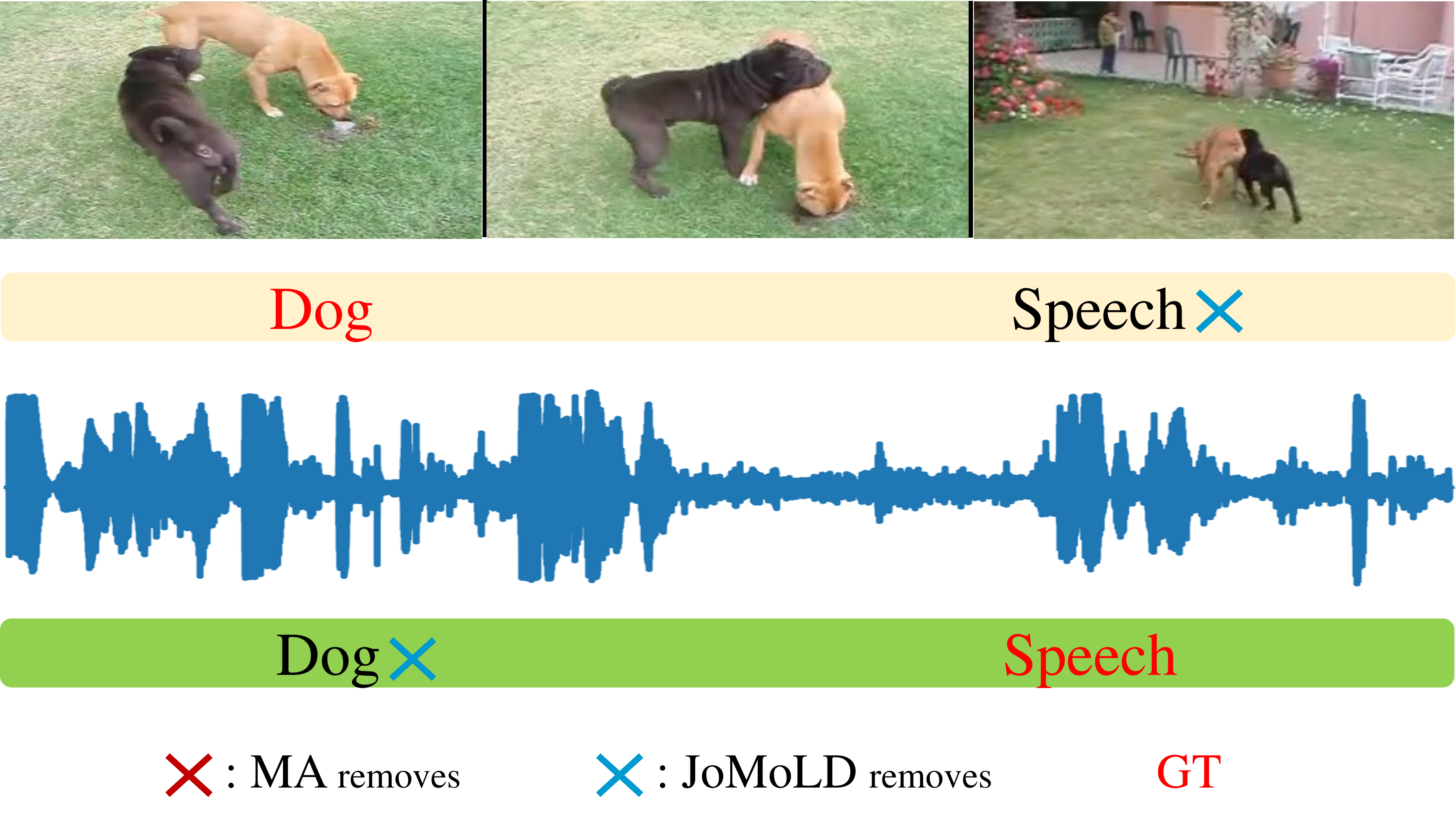}
        \caption{There are no ``Dog'' clues appear in audio track and no ``Speech'' clues appear in visual track. MA fails to identify the noise while JoMoLD correctly removes them.}
        \label{fig:compare2}
    \end{subfigure}
\caption{Comparisons of label denoising results between JoMoLD and MA.}
\label{fig:compare}
\end{figure}

\noindent\textbf{Label Denoising Visualization.} We visualize two cases of label denoising results of JoMoLD and MA~\cite{wu2021exploring} in Figure~\ref{fig:compare}.
The first case displays a motorcycle race.
The event ``Motorcycle'' can be perceived from both modalities, but ``Speech'' is from the off-screen audience.
In the second case, the event ``Dog'' only appears in the visual track and ``Speech'' in the audio track.
MA makes mistakes in both two cases.
When identifying noisy labels, MA exchanges the audio tracks of one video with another video whose label sets do not intersect. So the modality predictions of the newly assembled video are lowered when cross-attended to the unrelated modality.
While our method does not confuse the video content, and successfully removes noisy labels while retaining correct labels.

More visualizations are presented in the appendix.


\section{Conclusions}
In our work, we focus on weakly-supervised audio-visual video parsing task. 
We are committed to solving the modality-specific label noise issue, which degenerates parsing performance according to our analysis. We notice that the clean and noisy labels present different loss patterns, and an annotated event label would not be noise for both modalities. 
Thus we take the different loss levels of the two modalities as the noise and remove the noisy labels. Extensive experiments show that our Joint-Modal Label Denoising method selects modality-specific noise more accurately and improves performance over the state of the arts. 
As for the limitations, more large-scale datasets of the weakly-supervised AVVP task are expected to further validate our method in future work.

~\\
\noindent\textbf{Acknowledgements.}
This work is supported by National Natural Science Foundation of China  (No.62076119, No.61921006),  Program for Innovative Talents and Entrepreneur in Jiangsu Province, and Collaborative Innovation Center of Novel Software Technology and Industrialization.

\appendix

\section*{Appendix}
\title{Supplementary Materials \\
\normalsize Joint-Modal Label Denoising for Weakly-Supervised \\Audio-Visual Video Parsing}

\counterwithin{figure}{section}
\counterwithin{table}{section}

The appendix provides more visualizations and analyses to show deep insights into our method.
Sec.~\ref{sec:optimize} exhausts the details of optimizing the network after performing modality-specific label denoising in each training iteration. 
In Sec.~\ref{sec:experiments}, we conduct more ablation studies for our method.
To further compare JoMoLD with other methods (\ie, HAN~\cite{tian2020unified} and MA~\cite{wu2021exploring} ), Sec.~\ref{sec:case_study} provides more specific visualization cases.
\blfootnote{\small \Letter: Corresponding author.}

\section{Details of optimizing network $\mathcal{F}$}
\label{sec:optimize}
This section elaborates on the details of optimizing parsing network $\mathcal{F}$.

\noindent\textbf{Data Input.}
As described in Sec. 3.1 of the main paper, for a given video, pre-trained off-the-shelf networks extract segment-level audio and visual features ${\bf{f}}^a = \{f^a_1,...,f^a_t,...,f^v_T\}$, ${\bf{f}}^v = \{f^v_1,...,f^v_t,...,f^v_T\}$, where $t$ denotes the segment timestamp and $T$ represents the total number of segments. The fixed local features are fed into the network $\mathcal{F}$.

\noindent\textbf{Feature Aggregation.}
Previous work~\cite{tian2020unified} proves the significance of aggregating temporal context and leveraging the clues in different modalities. We define a function $Attn$ to represent the widely used attention mechanism:
\begin{equation}
\begin{aligned}
Attn(q,{\bf{K}},{\bf{V}}) = Softmax(\frac{q{\bf{K}}^T}{d}){\bf{V}},
\end{aligned}
\end{equation}
where d represents the dimension of vector $q$. Local features $f^a_t$ and $f^v_t$ are then enhanced by the following way:
\begin{equation}
\begin{aligned}
\hat{f}^a_t = f^a_t + Attn(f^a_t, {\bf{f}}^a, {\bf{f}}^a) + Attn(f^a_t, {\bf{f}}^v, {\bf{f}}^v) ,\\
\hat{f}^v_t = f^v_t + Attn(f^v_t, {\bf{f}}^v, {\bf{f}}^v) + Attn(f^v_t, {\bf{f}}^a, {\bf{f}}^a) ,\\
\end{aligned}
\end{equation}
The enhanced features $\hat{f}^a_t, \hat{f}^v_t$ are context-aware and have better capabilities of identifying the events occurred at $t$-segment.

\noindent\textbf{Model Output.} 
The audio-visual video parsing task predicts the event categories in each segment. In network $\mathcal{F}$, a shared fully-connected layer projects enhanced audio and visual features $\hat{f}^a_t$, $\hat{f}^v_t$ to label space. As a multi-label multi-class classification task, the sigmoid function is applied to further output the probabilities (between 0-1) for all event categories for each segment.
We express this process as the following equation:
\begin{equation}
\begin{aligned}
p^a_t = Sigmoid(FC(\hat{f}^a_t)), \quad
p^v_t = Sigmoid(FC(\hat{f}^v_t)),
\end{aligned}
\end{equation}
where $p^a_t, p^v_t \in (0,1)^{C}$.
During training, since only video-level labels are available, parsing network $\mathcal{F}$ adopts an attentive MMIL Pooling mechanism to obtain audio-level, visual-level and video-level event probability $p^a, p^v, p\in (0,1)^{C}$ by gathering weighted average of segment-level event probabilities:
\begin{equation}
\begin{aligned}
p^a[c] = \sum_{t=1}^T W_t^a[c]\,p_t^a[c], \quad
p^v[c] = \sum_{t=1}^T W_t^v[c]\,p_t^v[c], \\
p[c] = \sum_{t=1}^T W_t^{av}[0, c]W_t^a[c]\,p_t^a[c] + W_t^{av}[1, c]W_t^v[c]\,p_t^v[c],
\end{aligned}
\end{equation}
where $W_t^a, W_t^v \in (0,1)^C$ and $W_t^{av} \in (0,1)^{2\times C}$ are temporal and audio-visual attention weights respectively. $W^a=\{W^a_t\}_{t=1}^T, W^v=\{W^v_t\}_{t=1}^T\in(0,1)^{T\times C}$ are derived from applying learnable MLPs on $\hat{f}^a_t, \hat{f}^v_t$, and normalized by softmax function on temporal axis. And $W^{av}=\{W_t^{av}\}_{t=1}^T\in(0,1)^{T\times 2\times C}$ are derived from applying another learnable MLP layer on features $\hat{f}^a_t, \hat{f}^v_t$, then normalized on modality axis.

\noindent\textbf{Training Losses.}
We optimize the network in a batch manner. Let $B$ denote the batch size, and ${\bf{P}}^a$, ${\bf{P}}^v$, ${\bf{P}} \in (0,1)^{B \times C}$ represent the audio-level, visual-level and video-level event probabilities of a batch samples. The labels ${\bf{Y}}^a$, ${\bf{Y}}^v \in \{0,1\}^{B \times C}$ are the refined audio-level and visual-level labels obtained by modality-specific label denoising. ${\bf{Y}} \in \{0,1\}^{B \times C}$ represent the original video-level labels.
We can then optimize network $\mathcal{F}$ with audio-level loss $\mathcal{L}_a$, visual-level loss $\mathcal{L}_v$, and video-level loss $\mathcal{L}_s$ using binary cross-entropy loss:
\begin{equation}
\begin{aligned}
\mathcal{L} &=\mathcal{L}_a + \mathcal{L}_v + \mathcal{L}_s \\
&=-\frac{1}{B}\sum_{b=1}^B\sum_{c=1}^C {\bf{Y}}^a[b, c] log({\bf{P}}^a[b, c]) \\
&\quad-\frac{1}{B}\sum_{b=1}^B\sum_{c=1}^C {\bf{Y}}^v[b, c] log({\bf{P}}^v[b, c]) \\
&\quad-\frac{1}{B}\sum_{b=1}^B\sum_{c=1}^C {\bf{Y}}[b, c] log({\bf{P}}[b, c]).
\end{aligned}
\end{equation}

\section{More Ablation Studies}
\label{sec:experiments}

In this section, we explore more ablation studies to demonstrate the rationality of our method, which are not displayed in the main paper due to space limitation. Segment-level metrics are reported.

\textit{Study the Impact of Cross-modal Attention during \textbf{Calculating Forward Loss}}.
As stated in Sec. 3.3 of the main paper, we skip cross-modal attention when calculating forward loss in modality-specific label denoising. Cross-modal attention interferes with the event predictions of two modalities, and produces inaccurate modality-specific losses and further inaccurate noisy labels.
Results in Table~\ref{tab:skip_cm} quantify the effectiveness of removing cross-modal attention.

\textit{Study the Generality of JoMoLD Equipped with Other Baselines}. 
We mainly utilizes HAN~\cite{tian2020unified} as the backbone since it's a widely used baseline.
To further validate the generality, we change different backbones with JoMoLD. 
We modify two models from similar tasks as backbones, \textit{i.e.}, 1) AVE~\cite{tian2018audio}: audio-visual event localization task; 2) and AVSlowFast~\cite{xiao2020audiovisual}: audio-visual action recognition task. Table~\ref{tab:baseline} confirms that our approach works well for different baselines.

\textit{Study the Impact of Different Batch Sizes}.
We study the impact of different batch sizes on the final results in Table~\ref{tab:bs}. The results of the models trained on smaller sizes are slightly lower than that on batch size 128, which is the optimal setting in our experiments. 
Two reasons can explain this: 1) Noises might not be uniformly distributed in a smaller batch; 2) There are more round-off errors for smaller batch sizes when determining the number of noises in a batch. The results of model trained on a larger batch size fluctuate within acceptable ranges.

\begin{table}[htbp]
    \centering
    \caption{\textbf{Study the effectiveness of skipping cross-modal attention.}}
    \resizebox{0.9\linewidth}{!}{
        \begin{tabular}{c|ccccc}
            \toprule
            Forward Loss for Label Denoising & Audio & Visual & Audio-Visual & Type@AV & Event@AV \\ \hline
            \textit{Not Skip} cross-modal attention & 60.3 & 60.0 & 55.1 & 58.9 & 57.9 \\ 
            \textbf{\textit{Skip} cross-modal attention} & \textbf{61.3} & \textbf{63.8} & \textbf{57.2} & \textbf{60.8} & \textbf{59.9} \\ 
            \bottomrule
        \end{tabular}
    }
    \label{tab:skip_cm}
\end{table}

\begin{table}[htbp]\scriptsize
\centering
\caption{\textbf{Study the generality of JoMoLD on other backbones.}}
\resizebox{0.9\linewidth}{!}{
    \begin{tabular}{c|ccccc}
    \toprule
    Method & Audio & Visual & Audio-Visual & Type@AV & Event@AV \\ \hline
    AVE~\cite{tian2018audio} & 49.9 & 37.3 & 37.0 & 41.4 & 43.6 \\
    \textbf{AVE + JoMoLD} & \textbf{50.8} & \textbf{39.5} & \textbf{39.8} & \textbf{43.4} & \textbf{45.9} \\ \hline
    AVSlowFast~\cite{xiao2020audiovisual} & 47.2 & 50.8 & 39.8 & 45.9 & 47.0 \\
    \textbf{AVSlowFast} + \textbf{JoMoLD} & \textbf{48.9} & \textbf{60.1} & \textbf{43.7} & \textbf{50.9} & \textbf{52.1} \\
    \bottomrule
    \end{tabular}
}
\label{tab:baseline}
\end{table}

\begin{table}[H]\tiny
\centering
\caption{\textbf{Study the impact of different batch sizes.}}
\resizebox{0.9\linewidth}{!}{
    \begin{tabular}{c|ccccc}
    \toprule
    Batch size & Audio & Visual & Audio-Visual & Type@AV & Event@AV \\ \hline
    32 & 61.3 & 63.1 & 56.4 & 60.3 & 59.6 \\ 
    64 & 61.4 & 63.2 & 57.0 & 60.5 & 59.7 \\
    \textbf{128} & \textbf{61.3} & \textbf{63.8} & \textbf{57.2} & \textbf{60.8} & \textbf{59.9} \\
    256 & 61.4 & 63.6 & 57.1 & 60.8 & 59.5\\
    \bottomrule
    \end{tabular}
}
\label{tab:bs}
\end{table}


\section{Additional Qualitative Analyses}
\label{sec:case_study}

In this section, we present additional visualization cases to compare our JoMoLD with other methods on video parsing and label denoising.

\subsection{Visualizations of Video Parsing} 
We visualize the video parsing results of JoMoLD, HAN~\cite{tian2020unified} and MA~\cite{wu2021exploring} on different examples. ``GT'' denotes the ground truth annotations.
Each video lasts for 10 seconds.
Our method achieves more accurate parsing performance by acquiring reliable modality-specific supervision during training.

\begin{figure}[h]
  \centering
  \begin{subfigure}{\textwidth}
    \centering
      \includegraphics[width=12cm,height=5.5cm]{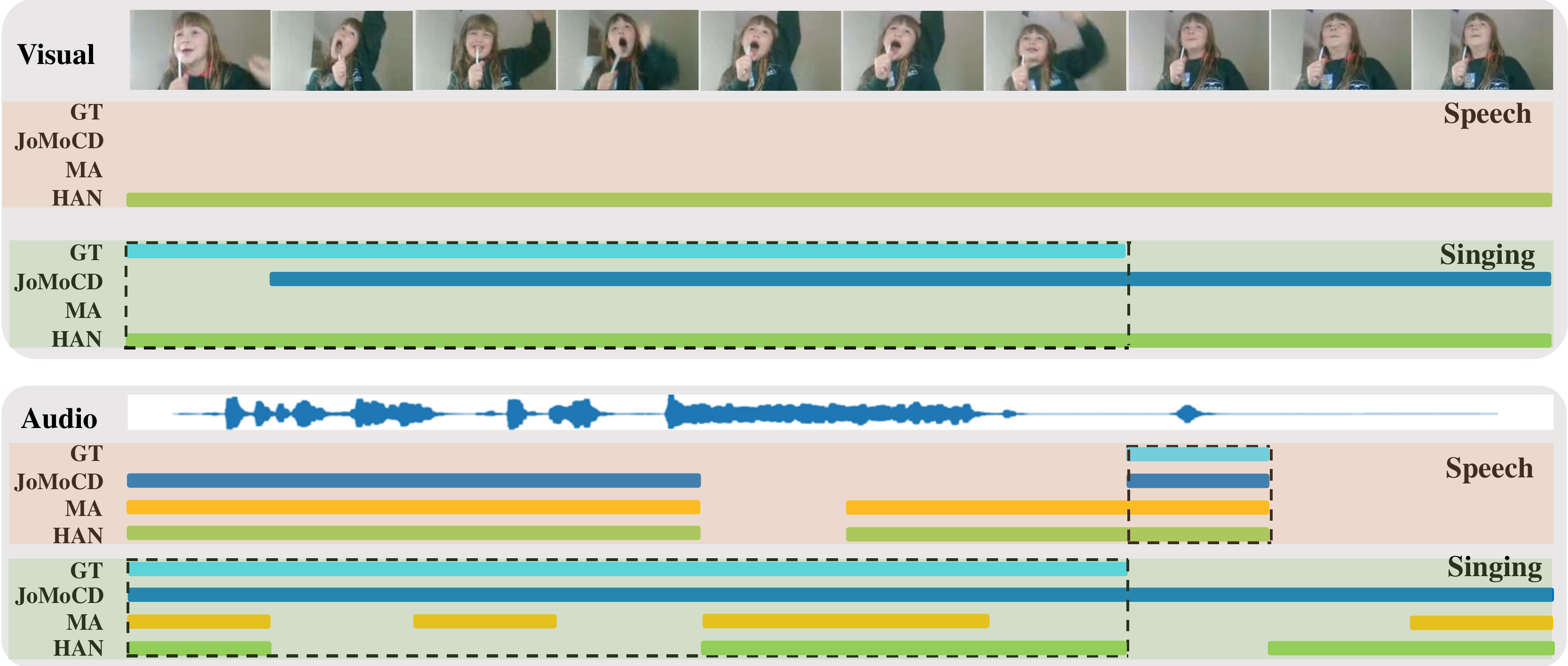}
      \caption{}
      \label{fig:parsing_69}
  \end{subfigure}
  \hfill
  \begin{subfigure}{\textwidth}
  \centering
      \includegraphics[width=12cm,height=5.8cm]{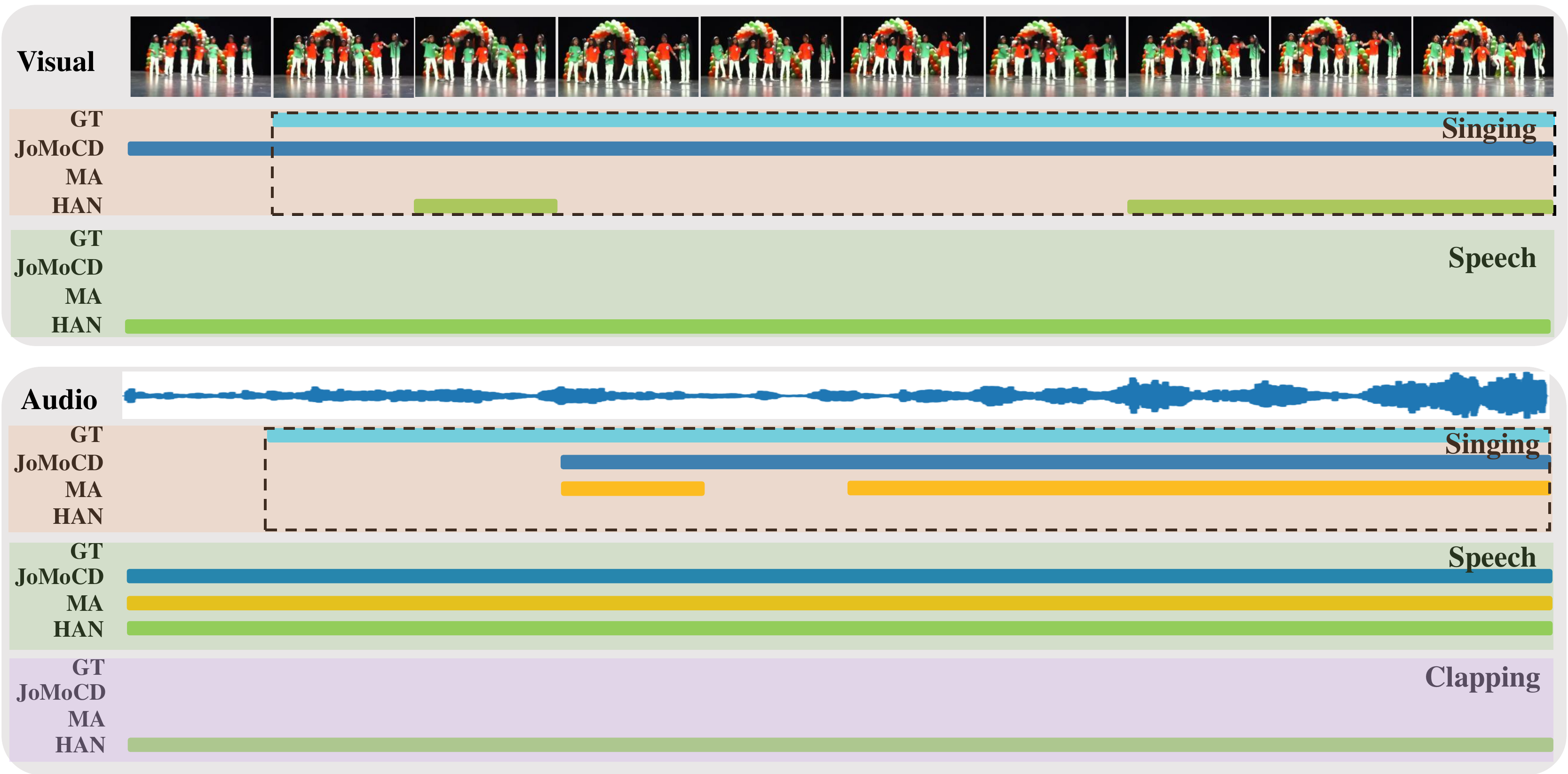}
      \caption{}
      \label{fig:parsing_837}
  \end{subfigure}
\end{figure}

\begin{figure}[H]\ContinuedFloat
  \centering
  \begin{subfigure}{\textwidth}
  \centering
      \includegraphics[width=12cm,height=5.5cm]{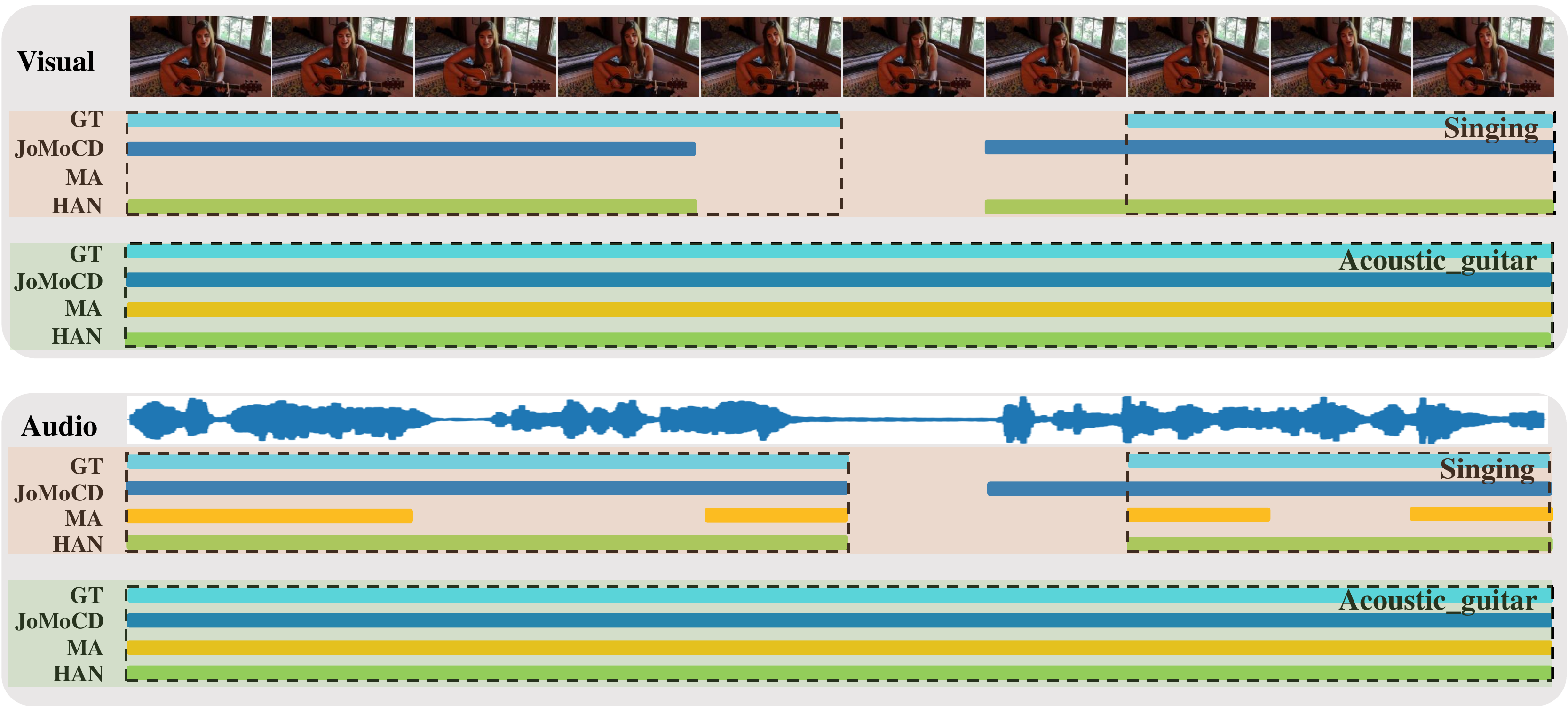}
      \caption{}
      \label{fig:parsing_1169}
  \end{subfigure}
  \begin{subfigure}{\textwidth}
  \centering
      \includegraphics[width=12cm,height=5.5cm]{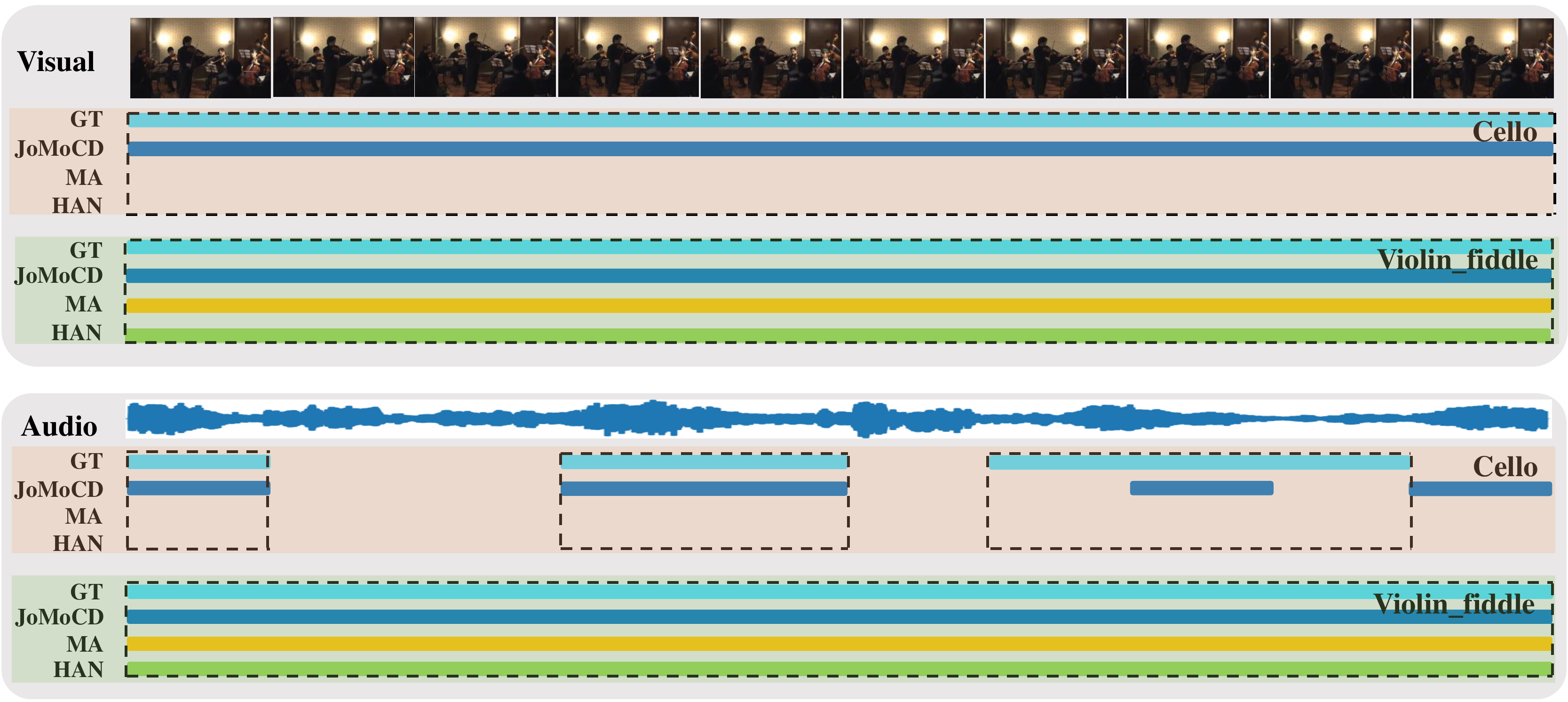}
      \caption{}
      \label{fig:parsing_496}
  \end{subfigure}
  \begin{subfigure}{\textwidth}
    \centering
      \includegraphics[width=12cm,height=5.5cm]{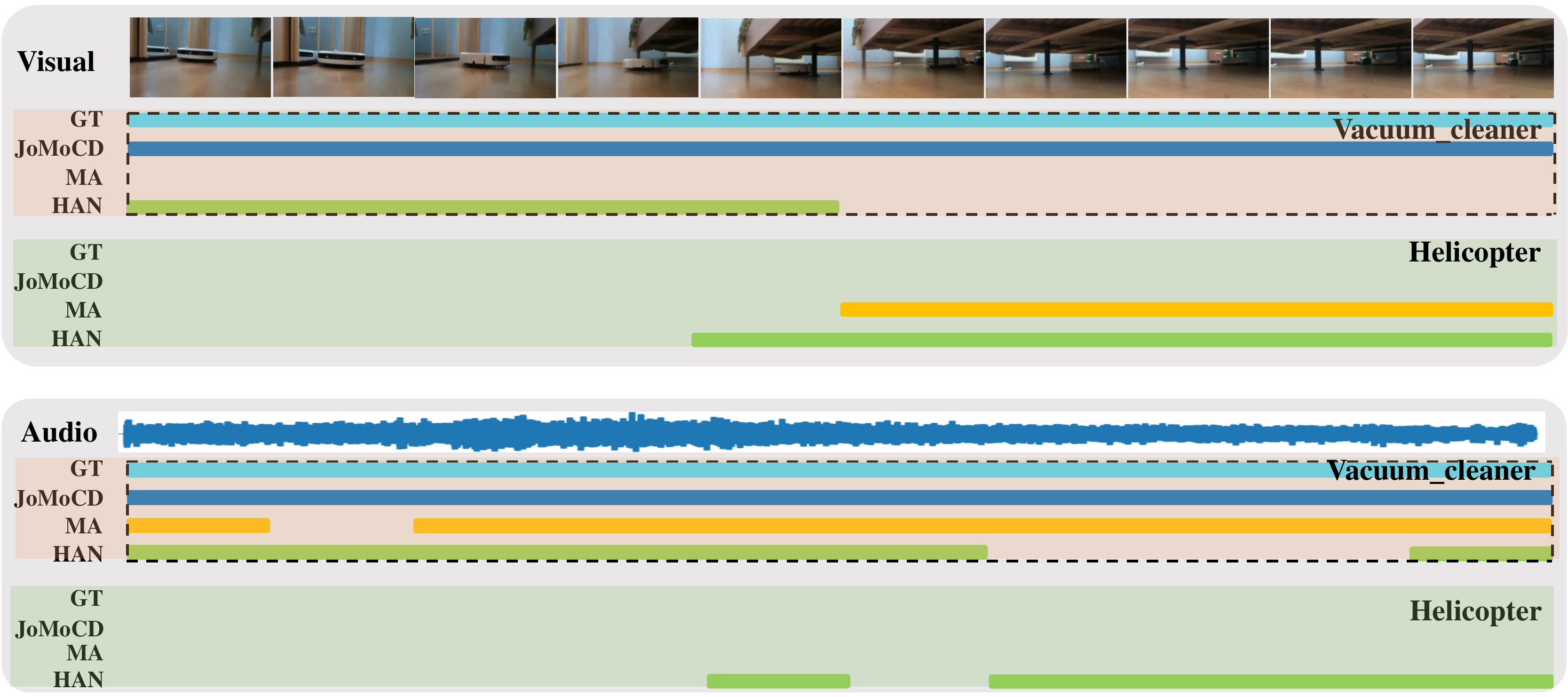}
      \caption{}
      \label{fig:parsing_472}
  \end{subfigure}
  \caption{\textbf{We visually compare JoMoLD with HAN, MA and ground truths.} In some easier examples, such as \ref{fig:parsing_496} and \ref{fig:parsing_472}, we achieve superior detection performance. In some difficult cases, the overall performance of JoMoLD is still better than HAN and MA.}
\end{figure}

\subsection{Visualization of Label Denoising}
In this section, we show that JoMoLD is superior to MA~\cite{wu2021exploring} in most cases when it comes to determining modality-specific noisy labels. 

On the one hand, as MA keeps cross-modal attention when performing modality-specific label denoising, the two modalities interfere with each other to cause inaccurate denoising results. While JoMoLD avoids cross-modal interference. 
On the other hand, MA adopts the naively trained baseline to determine the noisy labels. It trains the baseline with original videos but exchanges the audio tracks of two unrelated videos during label denoising, which leads to the gap between training and denoising. 
In contrast, there is no gap for JoMoLD, which consistently processes the original videos when training and denoising.
Meanwhile, JoMoLD adopts a dynamic manner to analyze the loss patterns of two modalities and remove noisy labels, which has a higher tolerance for denoising errors.

\begin{figure}
\centering
\begin{subfigure}{0.48\textwidth}
    \includegraphics[width=\textwidth]{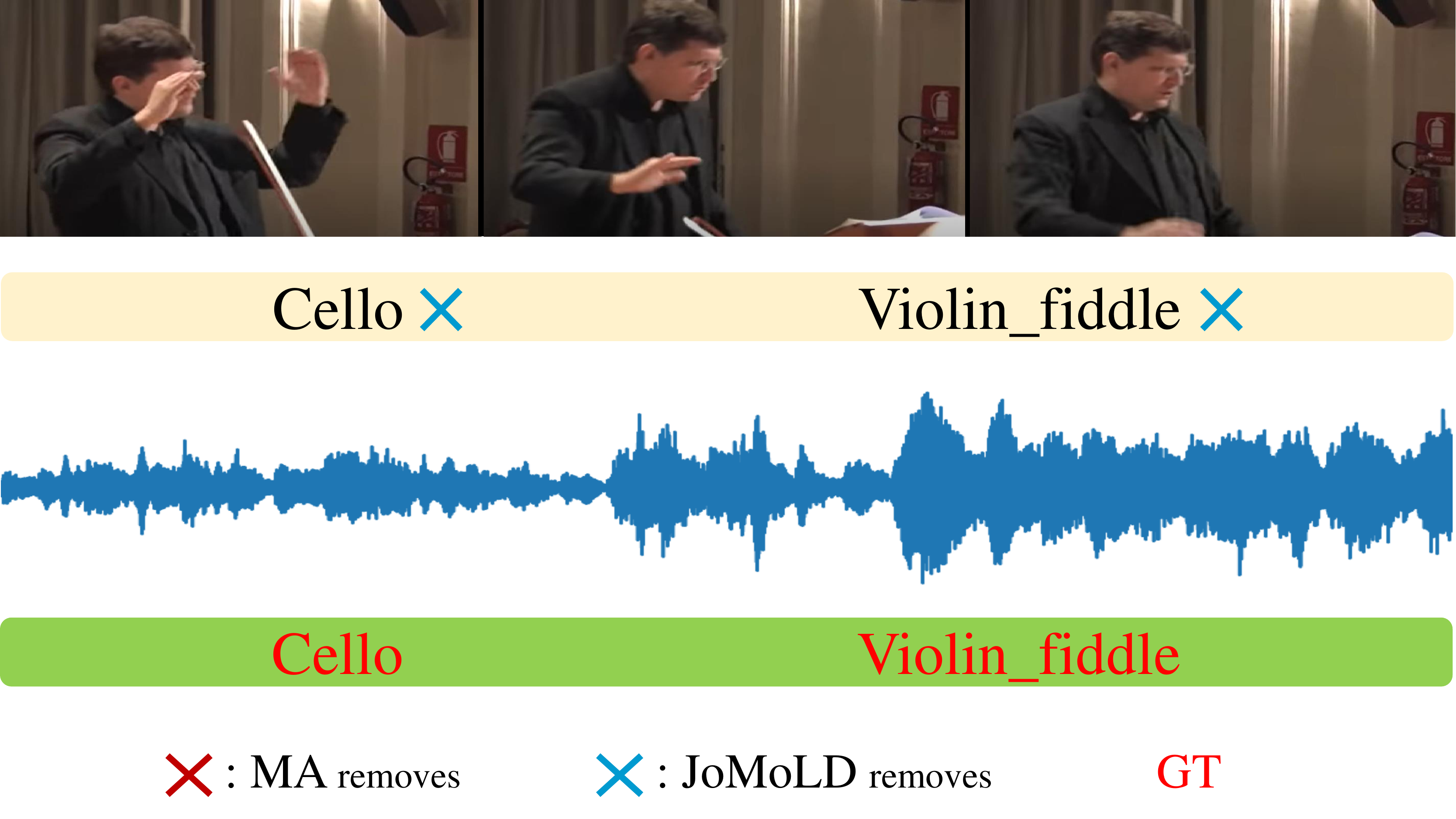}
    \caption{In this case, we show an example where MA fails to remove noisy labels, but our method correctly removes the noisy labels ``Cello'' and ``Violin$\_$fiddle'' for visual modality.}
    \label{fig:denoise_1511}
\end{subfigure}
\hfill
\begin{subfigure}{0.48\textwidth}
    \includegraphics[width=\textwidth]{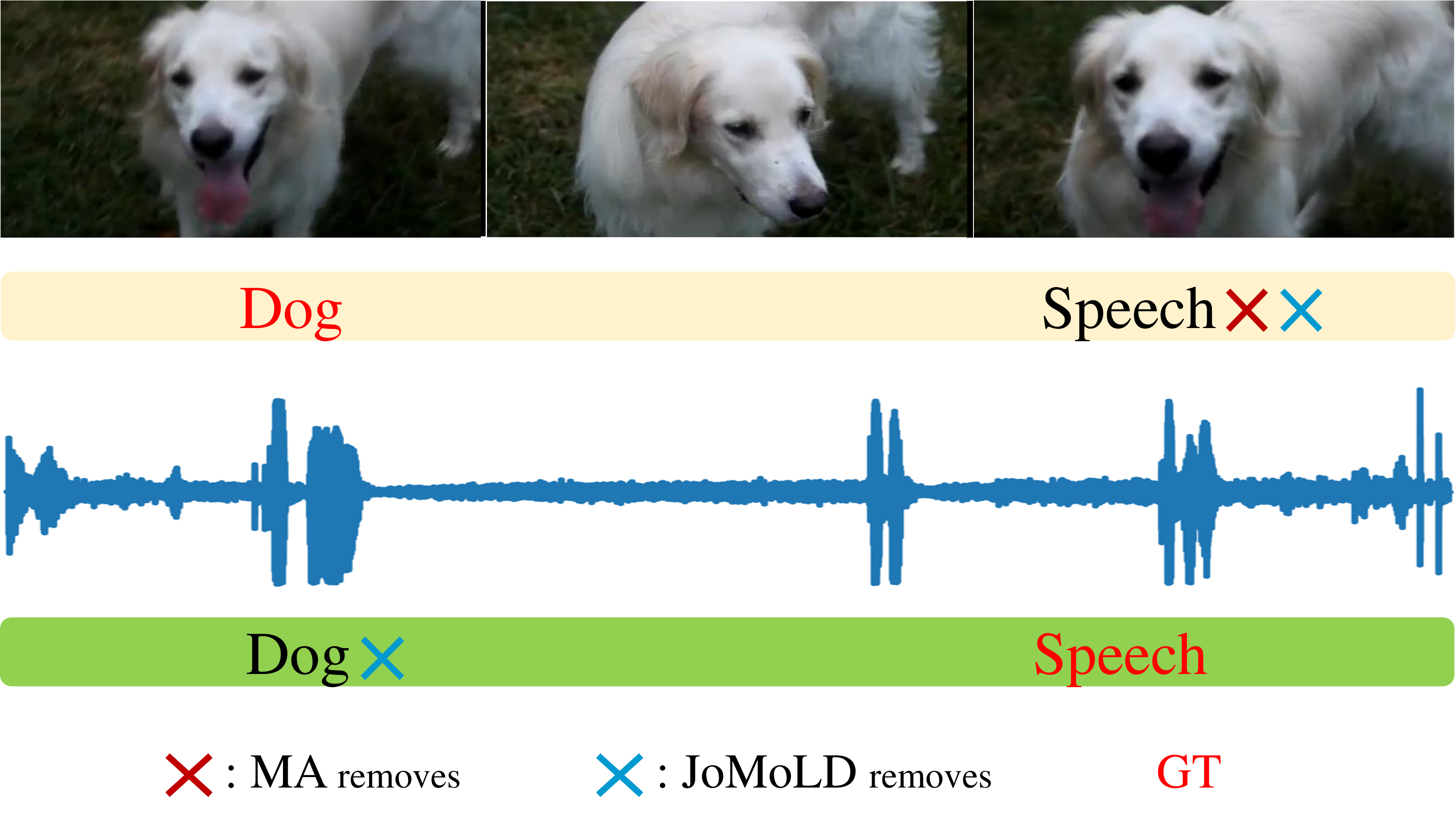}
    \caption{This case shows another example where our method correctly removes the audio noise MA does not filter out the noisy label.
    The dog do not bark, so ``Dog'' serves as audio noise.}
    \label{fig:denoise_9159}
\end{subfigure}
\quad
\hfill
\begin{subfigure}{0.48\textwidth}
    \includegraphics[width=\textwidth]{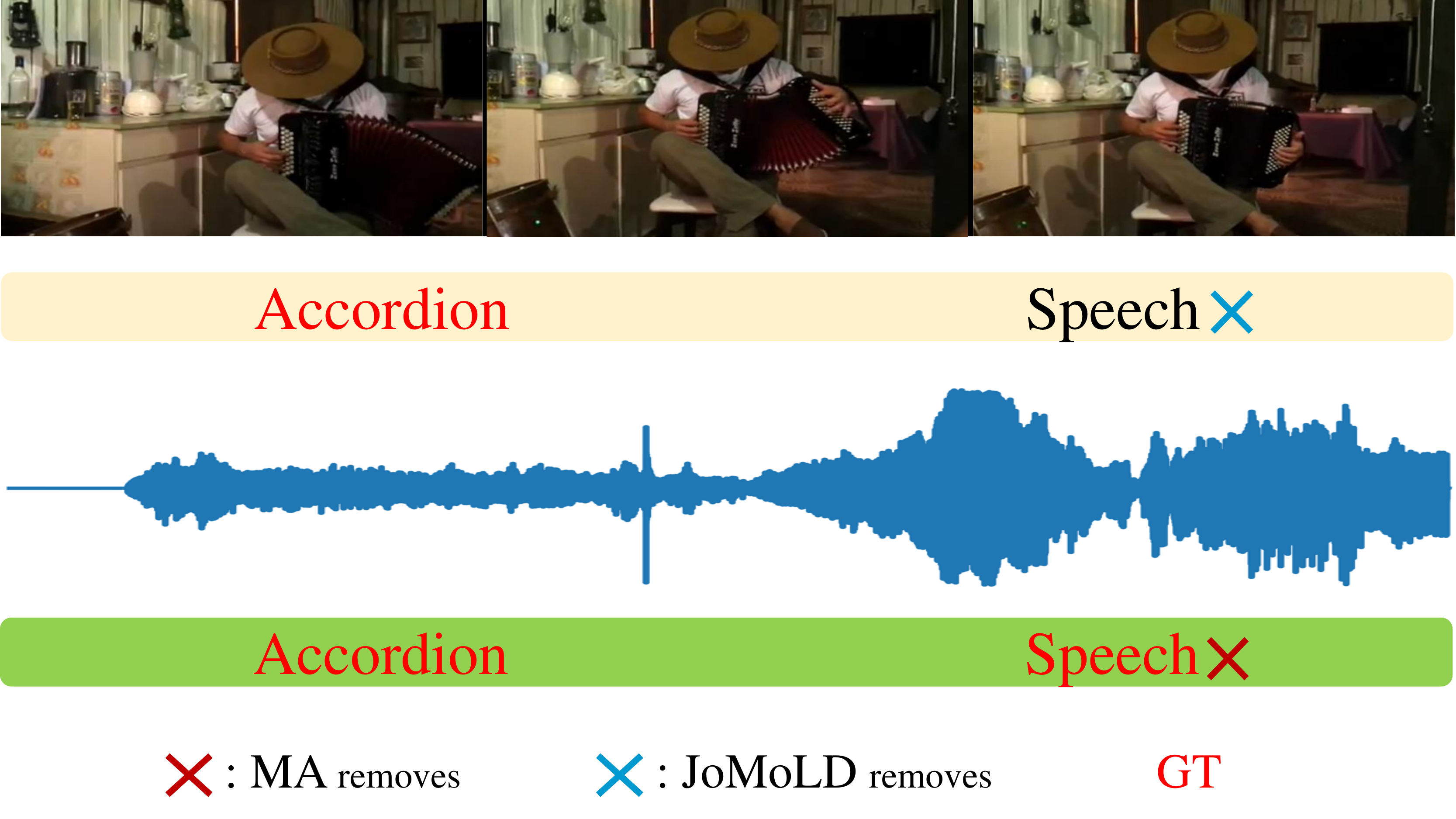}
    \caption{In the case, MA mistakenly removes the correct label but remains the noisy label.
    The person in the picture doesn't speak and another person is speaking off-screen. So ``Speech'' is a visual noise.}
    \label{fig:denoise_279}
\end{subfigure}
\hfill
\begin{subfigure}{0.48\textwidth}
    \includegraphics[width=\textwidth]{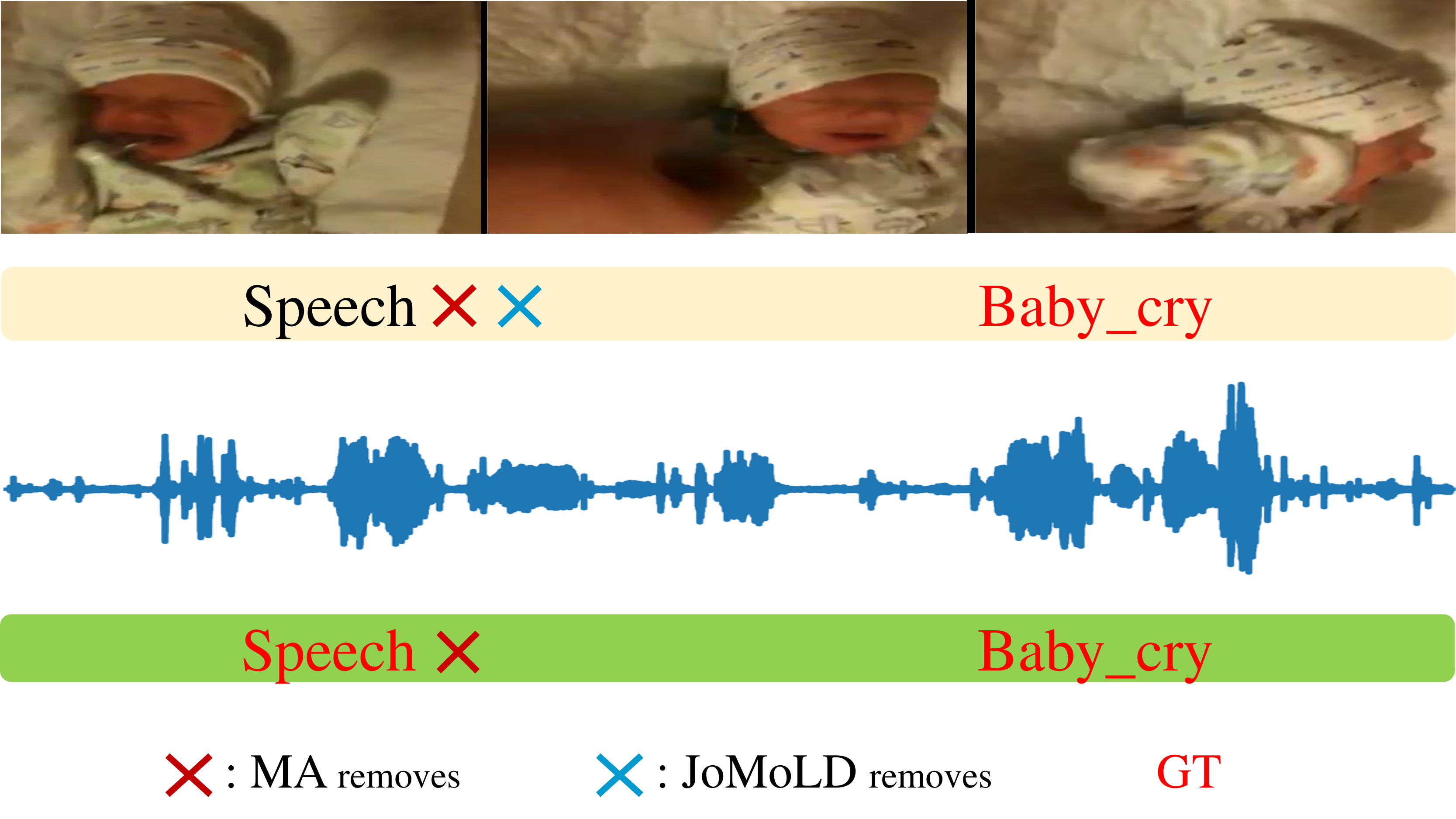}
    \caption{This case presents an example that our method correctly identifies the noisy modality but MA treats them both as noise. A parent is speaking off-screen so ``Speech'' is a visual noise.}
    \label{fig:denoise_1287}
\end{subfigure}
\caption{\textbf{Label denoising comparison between MA and JoMoLD.} We list four cases to illustrate different kinds of mistakes made by MA and avoided by our JoMoLD.}
\label{fig:denoise}
\end{figure}



\clearpage

\bibliographystyle{splncs04}
\bibliography{egbib}

\end{document}